%% file: PaperForReview.tex
\documentclass[10pt,twocolumn,letterpaper]{article}

\usepackage{wacv}              % To produce the CAMERA-READY version
\usepackage{graphicx}
\usepackage{amsmath}
\usepackage{amssymb}
\usepackage{booktabs}
\usepackage{graphicx}
\usepackage[]{threeparttable}
\usepackage{mathtools}
\usepackage{multirow}
\usepackage[aboveskip=0pt,belowskip=-3pt]{subcaption}

\usepackage[pagebackref,breaklinks,colorlinks]{hyperref}

\usepackage[capitalize]{cleveref}
\crefname{section}{Sec.}{Secs.}
\Crefname{section}{Section}{Sections}
\Crefname{table}{Table}{Tables}
\crefname{table}{Tab.}{Tabs.}

\input{tex/math_def}

\def\wacvPaperID{1830} % *** Enter the WACV Paper ID here
\def\confName{WACV}
\def\confYear{2024}

\begin{document}

\title{Learn to Unlearn for Deep Neural Networks:\\ Minimizing Unlearning Interference with Gradient Projection}

\author{Tuan Hoang 
\hspace{2em} Santu Rana \hspace{2em} Sunil Gupta \hspace{2em} Svetha Venkatesh \\
 A2I2, Deakin University\\
{\tt\small \{tuan.h;santu.rana;sunil.gupta;svetha.venkatesh\}@deakin.edu.au}
}
\maketitle

\input{tex/abstract}
\input{tex/introduction}
\input{tex/related_works}
\input{tex/proposed_method}

\input{tex/experiments}
\input{tex/conclusion}

\section*{Acknowledgement}
\vspace{-0.5em}
\noindent This research was partially funded by the Australian Government through the Australian Research Council (ARC). Prof Venkatesh is the recipient of an ARC Australian Laureate Fellowship (FL170100006).

%%%%%%%%% REFERENCES
{\small
\bibliographystyle{ieee_fullname}
\bibliography{egbib}
}

\newpage
%\title{- SUPPLEMENTARY - \\Learn to Unlearn for Deep Neural Networks:\\ Minimizing Unlearning Interference with Gradient Projection}

%\maketitle

\renewcommand{\thesection}{\Alph{section}}
\setcounter{section}{0}
\section{Experiments}

\subsection{Membership Inference Attacks}
\begin{figure*}[t]
\centering
\includegraphics[width=\textwidth]{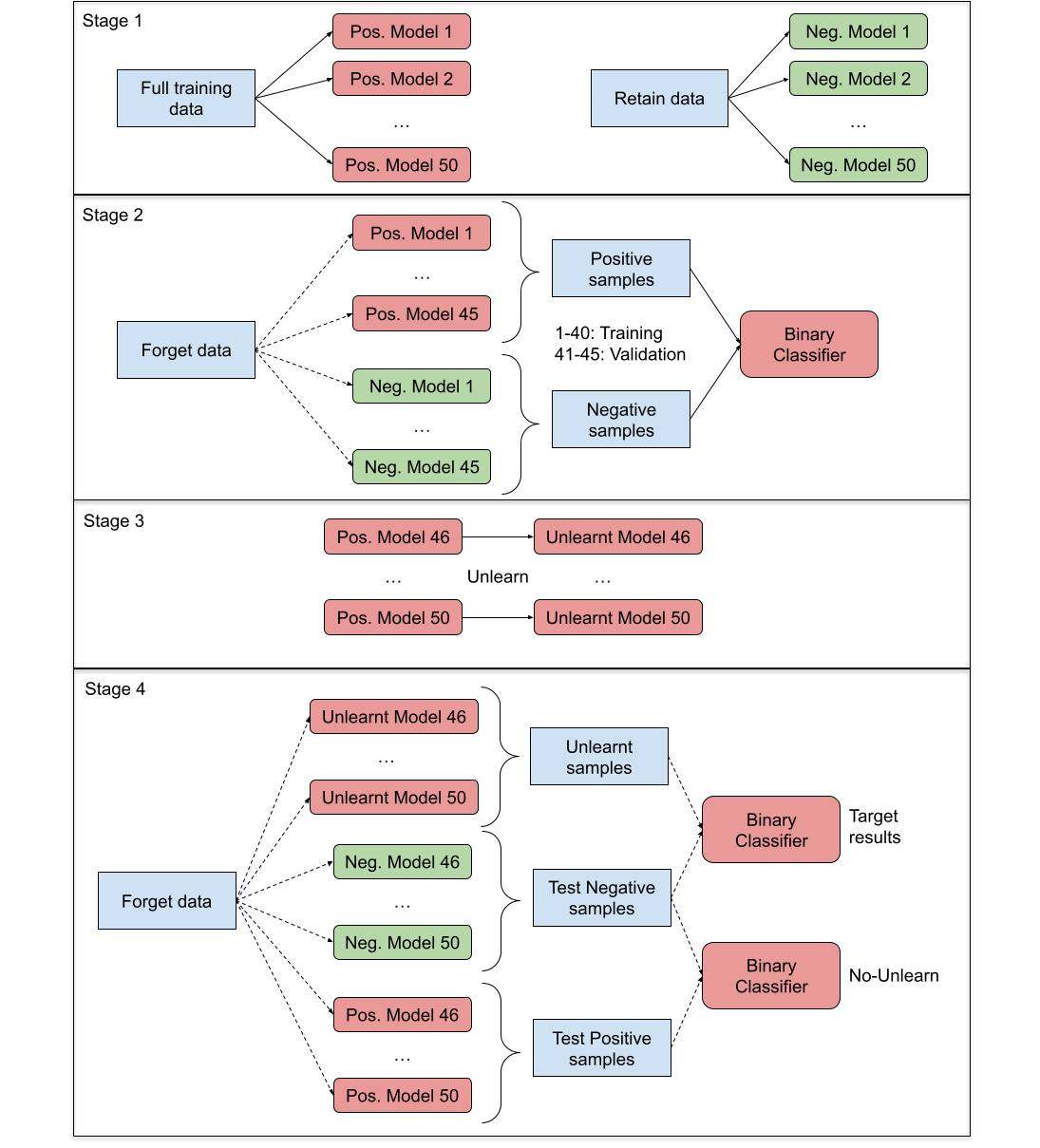}
\caption{The Membership Inference Attacks (MIA) experiment comprises 4 stages; \textbf{Stage 1}: We train 2 sets of models using the full training data (Positive models) and retrain data (Negative models). 
\textbf{Stage 2}: Given the forget data, we collect Positive samples and Negative samples (softmax outputs of the model) from Positive models and Negative models respectively. The Positive and Negative samples are used to train a binary classifier to detect whether a data point is used for training or not.
\textbf{Stage 3}: From positive models, we apply unlearning methods to obtain unlearnt models. 
\textbf{Stage 4}: Given the forget data, we then extract Test Positive, Test Negative and Unlearnt samples. Test Positive and Test Negative samples are combined to form a balanced no-unlearn test set. 
Unlearnt and Test Negative samples are combined to form a balance target test set.
Note: \textit{{Solid arrow lines}} indicate training process, \textit{{dash arrow lines}} indicate inferencing/testing process.
}
\label{fig:MIA_stages}
\end{figure*}

To conduct a membership inference attack (MIA), we train 2 sets of models: 50 original models with full training data and 50 retrained models with only the retaining dataset. We then treat the attack as a binary classification problem on model outputs of the forget dataset, where class \textbf{1} represents the samples seen during training (i.e., from original models) and class \textbf{0} represents the samples not seen during training (i.e., from the retrained model). 
For this classification, we use 40 models to extract a training set, 5 models to extract a validation set, and 5 models to extract a test set. The feature for MIA binary classifier comprises the softmax outputs of the models and the one-hot vector representing the sample class-label. We note that for class removal experiment, we remove the row corresponding to the forgetten classes in the MIA features.
Subsequently, we train an XGBoost classifier and fine-tune it to achieve the best F1 score on the validation set. Please refer to Figure \ref{fig:MIA_stages} for more details.  We visualize the Receiver Operating Characteristic (ROC) curve and report the Area Under Curve (AUC).
%
% in log scale as suggested in  \cite{carlini2022membership}.
The ROC curve illustrates the trade-off between the true-positive rate (TPR) and the false-positive rate (FPR). In their study \cite{carlini2022membership}, the authors focus on the extreme scenario of an exceedingly low FPR. They argue that de-identifying even a small number of users within a sensitive dataset is far more important than saying an average-case statement “most people are probably not contained in the sensitive dataset”. Nevertheless, in different contexts, such as when potential attackers seek to ascertain whether a user has contributed their data to a model's training, indicating interest in a particular application, the objective may change. Here, the goal is to identify as many data points (and their corresponding users) as possible. Therefore, potentially tolerating a higher FPR to achieve a higher TPR (i.e., Recall) is reasonable.

% Furthermore, it is worth noting that our assumption about the knowledge of attacker is quite strong; specifically, we assume that the attacker knows the model architecture, training methods, the full labelled training and forgetting dataset. Employing such a strong assumption simplifies the task of quantifying the extent to which information from the forgetting set still retains within the model. This might be more challenging in scenarios due to the biases caused by shallow models, shallow training-testing sets, etc.

\input{tex/ablation_studies}

\end{document}

%% file: tex/math_def.tex
\newcommand{\bw}{\mathbf{w}}

\newcommand{\bR}{\mathbf{R}}

\newcommand{\bsRl}{\boldsymbol{R}^l}
\newcommand{\bsRrl}{\boldsymbol{R}_r^l}
\newcommand{\bsRfl}{\boldsymbol{R}_f^l}

\newcommand{\bsUl}{\boldsymbol{U}^l}
\newcommand{\bsUrl}{\boldsymbol{U}_r^l}

\newcommand{\bsVl}{\boldsymbol{V}^l}

\newcommand{\bsSigmal}{\boldsymbol{\Sigma}^l}
\newcommand{\bsSigmarl}{\boldsymbol{\Sigma}_r^l}

\newcommand{\bsx}{\boldsymbol{x}}
\newcommand{\bsr}{\boldsymbol{r}}
\newcommand{\bsy}{\boldsymbol{y}}

\newcommand{\bsu}{{\boldsymbol{u}}}
\newcommand{\bsz}{{\boldsymbol{z}}}

\newcommand{\bsM}{{\boldsymbol{M}}}

\newcommand{\bbR}{\mathbb{R}}

\newcommand{\dset}{\mathcal{D}}

%% file: tex/abstract.tex
\begin{abstract}

Recent data-privacy laws have sparked interest in machine unlearning, which involves removing the effect of specific training samples from a learnt model as if they were never present in the original training dataset. The challenge of machine unlearning is to discard information about the ``forget'' data in the learnt model without altering the knowledge about the remaining dataset and to do so more efficiently than the naive retraining approach. To achieve this, we adopt a projected-gradient based learning method, named as Projected-Gradient Unlearning (PGU), in which the model takes steps in the orthogonal direction to the gradient subspaces deemed unimportant for the retaining dataset, so as to its knowledge is preserved. By utilizing Stochastic Gradient Descent (SGD) to update the model weights, our method can efficiently scale to any model and dataset size. We provide empirically evidence to demonstrate that our unlearning method can produce models that behave similar to models retrained from scratch across various metrics even when the training dataset is no longer accessible. Our code is available at \href{https://github.com/hnanhtuan/projected_gradient_unlearning}{https://github.com/hnanhtuan/projected\_gradient\_unlearning}.
\vspace{-2em}
\end{abstract}

%% file: tex/introduction.tex
\section{Introduction}

Deep learning has widely adopted across various fields, such as computer vision, natural language processing, and image/music/video generation. One of the reasons deep models excel is their ability to leverage vast quantities of data for training. However, machine learning models may unintentionally memorize their training data to a certain level, and recent work has shown that it is possible to derive meaningful information about individual training examples using only the parameters of a trained model \cite{7958568,10.1145/2810103.2813677,236216,melis2019exploiting, 7536387}. When the training data potentially contains privacy-sensitive user information, this creates significant challenges in regulating access to each user's data or enforcing personal data ownership, which the General Data Protection Regulation (GDPR) in the European Union \cite{GDPR} and the California Consumer Privacy Act (CCPA) \cite{CCPA} aim to address. Therefore, it becomes imperative to develop learning techniques that limit such memorization (such as Differential Privacy learning) or remove such model memorization when necessary (through Machine Unlearning), which is the main focus of this paper. Such a problem of machine unlearning can extends to the other applications such as de-poisoning, where we want to remove the effect of a subset of data previously used for training and later identified as malicious (e.g., anomalies) \cite{7163042} or biased \cite{10.1145/3457607} (denoted as ``poisoned samples''). After the unlearning process, the ML model ideally performs well as if the model has not been trained with the malicious/biased data.

When users invoke their \textit{``right to be forgotten'' (RTBF)}, it's crucial to ensure that the user data is ``unlearnt'' from the trained model. This means that any information derived from the requested-to-delete (forgetting) data should be removed from the model's knowledge.
The first plausible solution is to retrain the model from scratch without including the deleted data. However, this approach may be less practical due to its \textit{high computation, time}, and \textit{space costs}. 
% In general, the number of users who ask to delete their data is typically much lower than the number of remaining users, making it expensive to retrain the model by excluding the deleted data. 
Furthermore, this solution may sometimes require re-collection of training data to retrain the model as the training data may not be stored indefinitely due to privacy regulations \cite{GDPR}. For the same reason, it would be necessary for the unlearning method to work without requiring the training data.

Numerous methods \cite{ml_linear_filtration,phong_1,phong_2,fu2022knowledge,pmlr-v130-izzo21a,9857498,9577384,selective_forgetting,NTK_forget,L-CODEC,unlearn_features_labels,certified_linear_removal} have been proposed to facilitate unlearning. Many of these works \cite{selective_forgetting,NTK_forget,L-CODEC,unlearn_features_labels,certified_linear_removal} rely on the influence function \cite{influence_function}, that helps to estimate the influence of training data on the trained models, to find the update that reverse the effect of forgetting samples on this model. However, these works tend to be computationally intensive due to the Hessian estimation. As a result, it is difficult to achieve a significant runtime improvement for large models such as CNN over retraining.

In order to address the challenges of forgetting training knowledge with more computational efficiency, we first introduce a novel unlearning loss for classification which aims to reverse the original training process of the forgetting data.  
Then, inspired by recent works \cite{GPM,trgp}, we apply orthogonal gradient steps with respect to the core gradient subspace of the model weights for the retaining dataset. Specifically, we partition the entire gradient space of weights into two orthogonal subspaces: Core Gradient Space (CGS) and Residual Gradient Space (RGS) \cite{9605653}, where CGS contains information that needs to be preserved. This approach enables us to remove information related to the forgetting data from the trained model while inducing minimum interference with the retaining dataset, thereby avoiding catastrophic forgetting. We also propose a technique that can efficiently compute the CGS of the model weights for the retaining dataset from only the weight gradient space of the full training data, which is particularly useful when the full training dataset is no longer accessible.
% when the full training dataset is no longer accessible. This can be achieved by only caching  for full training data
% Importantly, we propose a novel unlearning loss for classification loss to reverse the training process of the forget samples.
% Existing approaches for unlearning [3–8] are inefficient in
% these cases, as they operate on data points only: First, a runtime
% improvement can hardly be obtained over retraining when the
% changes are not isolated and larger parts of the data need to
% be corrected. Second, removing multiple data points reduces
% the fidelity of the corrected model and thus is not a viable
% option in practical scenarios. Consequently, unlearning should
% not be limited to removing data points, but allow corrections at
% different granularity of the training data, such as fixing leaks
% in features and labels individually.

Our contributions can be summarized as follows:
\textbf{(i)} We propose an unlearning method for classification task with a novel unlearning loss function. The method utilizes gradient projection to remove information from a trained model with minimum interfering to important information of the retaining data, thereby preventing prevent catastrophic forgetting.
\textbf{(ii)} The proposed method only requires the forgetting data during the unlearning process and is applicable even when the training data is no longer accessible. \textbf{(iii)} In addition, our work can be applied to depoisoning application to eliminate harmful effects of poisoned training samples.
\textbf{(iv)} As our method employs gradient descent updates to unlearn the model, our method can be scaled effortlessly to any model and dataset.
% models of any size and datasets of any magnitude, which are feasible for training of the original model.
\textbf{(v)} Our experiment results on large scale models and datasets demonstrate that our unlearning method can produce models that behave similarly to models retrained from scratch across various metrics.

%% file: tex/related_works.tex
\section{Related works}

\subsection{Machine unlearning}
Earlier works on Machine Unlearning has been studied on the exact unlearning such as SVM \cite{5484614,10.1007/978-3-540-74690-4_22}, Naive Bayes classifiers, and $k$-means. However, these approaches are not suitable for CNNs, which are trained using stochastic gradient descent.

Yinzhi \textit{et al.} \cite{7163042} shows an efficient forgetting algorithm in the restricted setting of statistical query learning, where the learning algorithm cannot access individual samples. 
Bourtoule \textit{et al.} \cite{SISA} introduce the "sharded, isolated, sliced, and aggregated" (SISA) framework as a low-cost solution for knowledge removal, which involves the following three steps: (1) partition the complete training sample set into multiple disjoint shards, (2) train models independently on each of these shards, and (3) retrain the affected model upon receiving a request to unlearn a training point. However, this approach may incur a large storage overhead and its efficiency quickly deteriorates when multiple data points need to be removed.
% Recent works proposed several methods for ``one-shot forgetting'', those methods have some limitations.
Gou \textit{et al.} \cite{certified_linear_removal} propose a certified-removal mechanism, a very strong theoretical guarantee that an unlearned model is indistinguishable from a retrained model that never encountered the data in the first place. The method utilizes the influence function \cite{influence_function} for L2-regularized linear models that are trained using a differentiable convex loss function, such as logistic regressors. However, the method does not extend to DNN due to its strong convex assumption.
Golatkar \textit{el at.} \cite{selective_forgetting} propose a selective forgetting procedure for DNNs trained with SGD, using an information theoretic formulation and exploiting the stability of SGD. They propose a forgetting mechanism which involves a shift in weight space, and addition of noise to the weights to destroy information.
\cite{NTK_forget} proposes a scrubing method by adopting Neural Tangent Kernel (NTK), which posits that large networks during training evolve in the same way as their linear approximations \cite{NTK}. This allows the model information can be scrubbed in one step (i.e., ``one-shot forgetting'').
%The authors have pointed out an interesting finding that its the projection matrix on the orthogonal of the subspace spanned by the training samples 
However, this method faces the computational bottleneck as the size of NTK matrix grows exponentially with the number of training samples and classes. 
Based on the forgetting theory provided by Sekhari \textit{et. al.} \cite{remember_forget}, Mehta et. al. \cite{L-CODEC}  propose a measure for computing conditional independence called L-CODEC which identifies the Markov Blanket of parameters to be updated so that the method can be applied to large models. However, the method can still be computationally-intensive for a very deep and wide network.
Baumhauer \textit{et al.} \cite{ml_linear_filtration} introduce a forgetting method for logit-based classification models by applying a linear transformation to the output logits. However, this method only applies \textit{filtration} to the final linear layer, and other layer weights may still retain information. Therefore, this method may not effectively address data privacy concerns. Additionally, this approach has limited applicability, as it can only be used for class-wide data deletion. In the work by Kurmanji \textit{et al.} \cite{SCRUB}, they introduce a novel approach to unlearning through the optimization of a \textit{min-max} problem. Within this framework, the \textit{max} steps are designed to guide the student model to to distance its outputs from the teacher outputs on forgotten examples, effectively erasing forgotten information. While the \textit{min} steps align the student model outputs with the teacher model outputs on retained examples for restoring the model's performance on retained data, should it have been adversely affected by the \textit{max} steps.
Thudi \textit{et. al.} \cite{unrollingSGD} design of a new training objective penalty that limits the overall change in weights during SGD and as a result facilitates approximate unlearning. Zhang \textit{et. al.} \cite{PCMU}  initially establish a connection between randomized smoothing techniques for achieving certified robustness in classification tasks and randomized smoothing methods for certified machine unlearning with gradient quantization. Then based on that connection, they propose the concept of Prompt Certified Machine Unlearning (PCMU), which is built upon a foundation of randomized data smoothing and gradient quantization.

\subsection{Differential privacy}
% \vspace{-0.5em}
Differential Privacy (DP) \cite{private_data_analysis,Algorithmic_DP,Golatkar_2022_CVPR,andrew2021differentially,gopi2021numerical} offers a formal solution to address data privacy concerns and safeguard data ownership. Compared to unlearning, DP is more stringent as it restricts memorization and seeks to learn model parameters in a way that prevents retrieval of any information related to any training samples, while still achieving reasonable performance. On the other hand, unlearning merely aims to remove model information associated with a subset of training data after standard training, without expecting the model to perform well on those deleted samples. Due to its more rigorous requirements,  differential privacy for deep networks can be challenging to achieve and often results in significant accuracy losses. Therefore, when the requirement about data privacy is not excessively strict, machine unlearning may be a more suitable option.

\subsection{Membership Inference Attack}
% \vspace{-0.5em}
Membership Inference Attack (MIA) \cite{7958568,white_black_MI,choquette2021label,carlini2022membership} tries to determine if a particular data was used for training a model. This attack can serve as an effective means of evaluating the forgetting capacity of a model, especially when there are no or weak theoretical guarantees to quantify the remaining knowledge of forgetting data in model parameters.  If the attacker's predictions are comparable for both unlearned and retrained models, then it implies that the unlearned model has lost information that is specific to the forgetting data. In contrast, if the attacker displays a greater degree of confidence in predicting a forgetting sample from an unlearned model as opposed to a retrained one, it could indicate ineffective unlearning. Conversely, a lower confidence than random chances in predicting from an unlearned model could lead to the Streisand Effect.
% A higher accuracy value indicates incorrect (or no) unlearning, while a lower value may result in the Streisand Effect.
% As a unlearned model is deemed to have forgotten a sample only if an attacker predict its sample membership the same as a retrained model. 

%% file: tex/proposed_method.tex
\section{Proposed method}

\subsection{Problem statement}
% \vspace{-0.5em}
Let $\dset=\{\bsx_i, \bsy_i\}_{i=1}^N$ be a fixed training dataset and $f_\bw(\bsx)$ be a parametric function (model), for instance a CNN, with parameters $\bw$ (weights) trained on $\dset$.
Let $\dset_f \subset \dset$ be a subset of the training data, whose information we want to remove from the model $f_\bw(x)$ (i.e., \textit{forgetting dataset}), and let $\dset_r$ be the complement of $\dset_f$ (i.e., $\dset_f \cup \dset_r=\dset$ and $\dset_f \cap \dset_r=\emptyset$ ), whose information we want to retain (i.e., \textit{retaining dataset}).

% \begin{equation}
% \begin{split}
%     \MI{\dis_f}{F_{\bar{\theta}}(\bx)}\le\int\int P\big(\dis_f|F_{\bar{\theta}}(\bx)\big)&P\big(F_{\bar{\theta}}(\bx)\big) \frac{P\big(\dis_f|F_{\bar{\theta}}(\bx)\big)}{Q\big(F_{\bar{\theta}}(\bx)\big)} \\
%     &=\bbE_{P\big(F_{\bar{\theta}}(\bx)\big)}\Big[KL\Big(P\big(\dis_f|F_{\bar{\theta}}(\bx)\big)||Q\big(F_{\bar{\theta}}(\bx)\big)\Big)\Big]
% \end{split}
% \end{equation}

% \begin{equation}
%     \MI{\bx}{\bz}\le \int\int P(\bz|\bx)P(\bx)\log\frac{P(\bz|\bx)}{Q(\bx)}=\bbE_{\bx\sim P(\bx)}\Big[KL\big(P(\bz|\bx)||Q(\bz)\big)\Big]
% \end{equation}
% Minimizing the mutual information between the forgetting data points $\bx$ and their corresponding outputs $\bz$ can be obtained by enforce the outputs $\bz$ to a pre-defined distribution $Q(\bz)$. Choosing any arbitrary distribution $Q(\bz)$ (e.g., a standard Gaussian distribution) can help to minize the mututal information; however, it can potentially lead to the “Streisand Effect” when the distribution of the outputs $\bz$ is obviously different from the distribution of other non-training samples (i.e., target distribution).

\subsection{Unlearning}
\label{sec:proposed_method}
% \vspace{-0.5em}

Our approach leverages the property that stochastic gradient descent (SGD) updates lie in the span of input data points \cite{zhang2017understanding}. Inspired by the works of Schulman \textit{et al.} (2015) \cite{GPM} and Lin \textit{et al.} \cite{trgp}, we have developed a method to selectively forget parts of a dataset by applying gradient updates orthogonal to the Core Gradient Space (CGS) \cite{9605653} of model weights computed using the retaining data. This approach allows the model to update its weights in a way that discards information about the forgetting data while preserving the knowledge learnt from the retaining data.

\subsubsection{Loss function:}
We first propose the loss function that reverses the learning process as follows:
\begin{equation}
\label{eq:unlearning_loss}
  % \mathcal{L}= \left(\sum_{i\in \dset_f} \sum_{c=1}^C -y_{i, c}\log (1 - p_{i, c} + \epsilon) - \lambda p_{i, c}\log(p_{i, c})\right) + \bb^\top \bw,
    \mathcal{L}= \sum_{i\in \dset_f} \sum_{c=1}^C \left(-y_{i, c}\log (1 - p_{i, c} + \epsilon) - \lambda p_{i, c}\log(p_{i, c})\right),
\end{equation}
where $p_{i, c}=\frac{\exp(z_{i,c})}{\sum_{j=1}^C \exp(z_{i,j})}$; $z_{i,c}$ is the $c$-th element of $\bsz_i=f_\bw(\bsx_i)$, and similarly $y_{i,c}$ is the $c$-th element of $\bsy_i$.

The first-term is ``\textit{reverse}'' cross-entropy loss which tries to \textit{minimize} the predicted probability of the true-label class. $\epsilon$ is a small constant value to avoid exponentially large gradients at the beginning of training when the scores can be high (i.e., $p_{i,c}\approx 1$). The second-term attempts to \textit{maximize} ($\lambda > 0$) the entropy of the model outputs, thereby making equal confidence scores  for all classes. This term is helpful in removing information from the forgetting data that could be useful in multiple classes (e.g., bird and airplane might share similar background information). In other words, the model also become less focused on features that are correlated with the forgetting labels.
Moreover, the second term also prevents the confidence scores of the forgetting data from going arbitrarily close to $0$, which could otherwise result in abnormal confidence scores for the forgetting data. 
% We finally perturb the empirical risk by a random linear term\cite{chaudhuri11a} in the last term. This term is to avoid information leakage due to the direction of the gradient \cite{certified_linear_removal}.
  
For the de-poisoning application, we not only want the model to unlearn the features (which could be noise) that are correlated with the "poisoned" labels, but we also expect the model to correctly re-classify these poisoned samples. Therefore, we use $\lambda < 0$ to minimize the entropy of the model outputs. This means that we want the model to assign a high confidence score to a class other than the poisoned labels. 
% In essence, we want to boost the model's confidence in correctly identifying the class of the affected samples.

\subsubsection{Core Gradient Space (CGS) construction:}
When the model training on full dataset is finished, we then compute the basis vectors and eigen-values of the gradient space for full dataset. Specifically, for each convolutional or linear layer $l$, we take forward pass for a training sample $\bsx_i$ to obtain the outputs $\bsz_i^{l-1}$ of $(l-1)$-th layer ($\bsx_i^l\equiv \bsz_i^{l-1}$ as input of $l$-th layer). For a convolution layer, we extract an input feature vector $\bsr_i^l$ by taking a patch vector from the 3-dimensional feature map $\bsz_i^{l-1}$; while for a linear layer, the input feature vector is simply $\bsr_i^l=\bsz_i^{l-1}$. We then 
concatenate all $d$-dimension input feature vectors along the column to construct an input-representation matrix $\bR^l=[\bsr_{1}^l, \bsr_{2}^l, \cdots, \bsr_{n^\ast}^l]\in\bbR^{d\times n^\ast}$. 
% Next, we perform SVD on $\bsRl=\bsUl\bsSigmal(\bsVl)^\top$. 
Next, we can compute the basis vectors $\bsUl$ and eigen-values $\bsSigmal$ by using SVD as follows:
\begin{equation}
    \bsRl(\bsRl)^\top=\sum_{\forall i}\bsRl_i(\bsRl_i)^\top=\bsUl(\bsSigmal)^2(\bsUl)^\top,
\vspace{-0.5em}
\end{equation}
where $\bsRl_i\in\bbR^{d\times m}$ is a subset of $\bsRl$ including $m$ input representations; hence, $\bsRl(\bsRl)^\top$ can be computed in mini-batch manner. As a result, $\bsUl$ and $\bsSigmal$ can be computed very efficiently (noting that $d$ is usually small, e.g., $< 1000$).

Whenever a data deletion request is received for $\dset_f$, we can compute the basis vectors of weight gradient space for the retaining dataset $\dset_r$ as follows (note that $\bsRl=[\bsRrl, \bsRfl]$):
\begin{equation}
\begin{split}
    \bsRrl(\bsRrl)^\top &= \bsRl(\bsRl)^\top - \bsRfl(\bsRfl)^\top \\
    \bsUrl(\bsSigmarl)^2(\bsUrl)^\top &=\bsUl (\bsSigmal)^2(\bsUl)^\top - \bsRfl(\bsRfl)^\top.
\end{split}
\end{equation}
We pre-compute and cache the basis vectors $\bsUl$ and eigen-values $\bsSigmal$ of the input representations of each layer of the full training set, that allows us to efficiently compute the basis vectors and eigen-values of the retaining data, i.e., $\bsUrl$ and $\bsSigmarl$, when the forgetting dataset is given. Importantly, our method does not require the training data\footnote{When $\bsVl$ is discarded, it is impossible to reconstruct $\bsRl$.}; so it is applicable even when the training data is not accessible anymore (except for the forgetting data $\dset_f$ that should be provided again by the users requesting data deletion). Our method can avoid the privacy concern over storing training data.

Next, we can obtain the Core Gradient space, $CGS=span\{\bsu_{r,1}^l, \bsu_{r,2}^l, \cdots, \bsu_{r,k}^l\}$, spanned by the first $k$ vectors of $\bsUrl$, where $k$ satisfies the following criteria for a given threshold $\gamma^l$: $\sum_{i=1}^k\sigma_{r,i}^l \ge \gamma^l\sum_{i=1}^d\sigma_{r,i}^l$ with $\bsSigmarl=\text{diag}([\sigma_{r,1}^l,\cdots, \sigma_{r,d}^l])$.
CGS can be presented in matrix format as $\bsM=[\bsu_{r,1}^l, \bsu_{r,2}^l, \cdots, \bsu_{r,k}^l]$.

\input{tex/exp_data_removal_table}

\subsubsection{Gradient update} Given the forgetting dataset $\dset_f$ and the loss function \ref{eq:unlearning_loss}, we can compute the gradient $\nabla_{\bw^l}L$. However, before applying the gradient step, the gradient $\nabla_{\bw^l}L$ are first projected onto the CGS and then projected components are subtracted out from the gradient so that the remaining gradient components lie in the space orthogonal to CGS. The gradients are processed as follow:
\begin{equation}
\label{eq:gradient_projection}
    \nabla_{\bw^l}L_{\perp}=\nabla_{\bw^l}L - \big(\nabla_{\bw^l}L\big) \bsM^l(\bsM^l)^\top.
\end{equation}
% For the de-poisoning application, since the unlearning process is strongly correlated with the learning of retain dataset (as we also aim to correctly re-classify poisoned samples),  
% Note that instead of using the naive gradient projection (GPM) \ref{eq:gradient_projection} \cite{GPM}, we can adopt the Trust Region Gradient Projection (TRGP) method \cite{trgp}. 
% We found that TRGP is more effective for the de-poisonining application since we also aim to re-classify poisoned samples, which means that the unlearning of forget dataset is strongly correlated with the learning of retain dataset. While we do not see any significant different in various metrics for the unlearning setting between TRGP and GPM. This is because the unlearning of forget dataset does not have strong correlation with the learning of retain dataset.
For the context of de-poisoning application, since the unlearning of the forgetting dataset (i.e., the poisoned dataset) is strongly correlated with the learning of the retaining dataset (as we also aim to re-classify poisoned data), simply applying the naive orthogonal gradient projection (Eq. \ref{eq:gradient_projection}) will compromise the unlearning process, as noted in \cite{trgp}. We address this issue by adopting the Trust Region Gradient Projection (TRGP) method \cite{trgp}\footnote{We refer readers to \cite{trgp} for the details of TRGP method.} and using the CGS of the retaining dataset as the trust region. We apply TRGP only to the last linear layer in the de-poisoning application, while Eq. \ref{eq:gradient_projection} is used in all other layers.

\vspace{0.5em}
As with the training of a classification model, there is a potential for overfitting to occur during the unlearning process. To mitigate this issue, we adopt  early stopping. Specifically, we compare the accuracy of the validation dataset to that of the forgetting dataset or to the random performance. By doing so, we can determine when the model starts to overfit to the unlearning and stop it early. 
% Alternatively, we can stop the unlearning early if the performance of the validation set drops more than a pre-defined threshold, e.g., $0.5\%$.

\subsubsection{Incremental Unlearning} 
\label{sec:incremental_unlearning}
\vspace{-0.5em}
It is common for user data to be deleted multiple times during the life cycle of a model. Therefore, it is essential for the unlearning method to support incremental unlearning. This means that the method should be able to forget multiple batches of data one-by-one, allowing the model to adjust its weights accordingly as more training data is removed.
\begin{equation}
\begin{split}
    &\bw_f^l = \bw_o^l +\Delta \bw_{o\to f}^l \\
    \Rightarrow~~  &\bw_f^l \bsr_i^l = \bw_o^l \bsr_i^l + \Delta \bw_{o\to f}^l \bsr_i^l \\
    \Rightarrow~~  &\bw_f^l \bsr_i^l \approx \bw_o^l \bsr_i^l.
\end{split}
\vspace{-0.5em}
\end{equation}
Here, we denote by $\bw_o^l$ the weight of the originally  model trained on $\dset$ and by $\Delta \bw_{o\to f}^l$ the update of weight from the original model to the unlearned model.
Since our weight updates lie in the CGS of the retaining data, it implies that $\Delta \bw_{o\to f}^l \bsr_i^l\approx 0 $.
As a result, the outputs of the unlearned model and those of the original model will be almost identical for the retaining dataset.
Therefore, given the retaining dataset at any time, the basis vectors $\bsUrl$ and eigen-values $\bsSigmarl$ of the original model and unlearned model are the same, allowing us to easily re-compute the basis vectors whenever the retaining dataset changes (i.e., more data points are removed). 
% Similar to \cite{GPM}, the hyper-parameter threshold $\gamma^l$ controls the stability-plasticity, in which stability represents the ability to preserve retaining data knowledge, while plasticity represents the ability to remove forgetting data knowledge.
% However, we note that using a small CGS or training the model for too long can affect the performance of the unlearned model on the retaining dataset, as $\Delta \bw_{o\to f}^l \bsr_i^l\neq 0$.

%% file: tex/exp_data_removal_table.tex
\begin{table*}[t]
\caption{The experiment results for various readout functions for unlearning 500 samples of Class 0 of CIFAR10 using AllCNN model.}
\label{tb:compare}
\vspace{-0.5em}
\def\arraystretch{1.1}
% \resizebox{1.0\textwidth}{!}
% \vspace{-1pt}
\centering
{
\begin{threeparttable}
\begin{tabular}{|l|c|c|c|c|c|}
\hline
Method & Retrained & EU$k$ & NTK &  SCRUB & \textbf{PGU}  \\ \hline \hline
$\dset_{r}$ error & 0.04$\pm$0.01 & 0.06$\pm$0.01 & 3.49$\pm$0.98 &  0.01$\pm$0.00 & 0.05$\pm$0.01 \\ 
$\dset_{f}$ error & 9.28$\pm$1.25 & 0.12$\pm$0.16 & 11.05$\pm$1.83 &  10.45$\pm$1.61 & 9.92$\pm$0.45   \\ 
$\dset_{\text{test}}$ error & 9.78$\pm$0.19 & 10.13$\pm$0.15 & 12.68$\pm$0.74 &  10.26$\pm$0.44 & 9.86$\pm$0.15  \\
Time (s) &  1021$\pm$18 & 821$\pm$22 & --$^\ast$ & 321$\pm$43 & 88.45$\pm$4.12 \\ \hline
\end{tabular}
\begin{tablenotes}
\footnotesize
\item $^\ast$ We do not report time for NTK since we can only samples a small subset of retaining dataset.
\end{tablenotes}
\end{threeparttable}
}
\end{table*}

\begin{figure*}[t]
\centering
\begin{subfigure}[b]{\textwidth}
\begin{subfigure}[b]{0.24\textwidth}
\centering
\includegraphics[width=\textwidth]{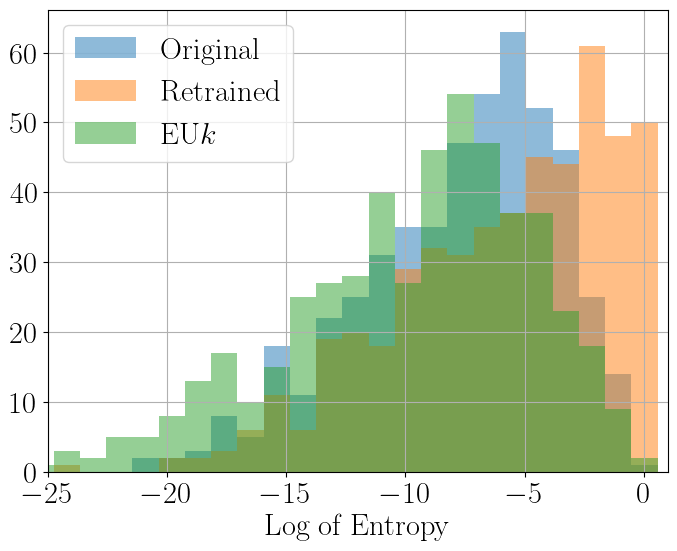}
% \caption{EU$k$}
\end{subfigure}
\begin{subfigure}[b]{0.24\textwidth}
\centering
\includegraphics[width=\textwidth]{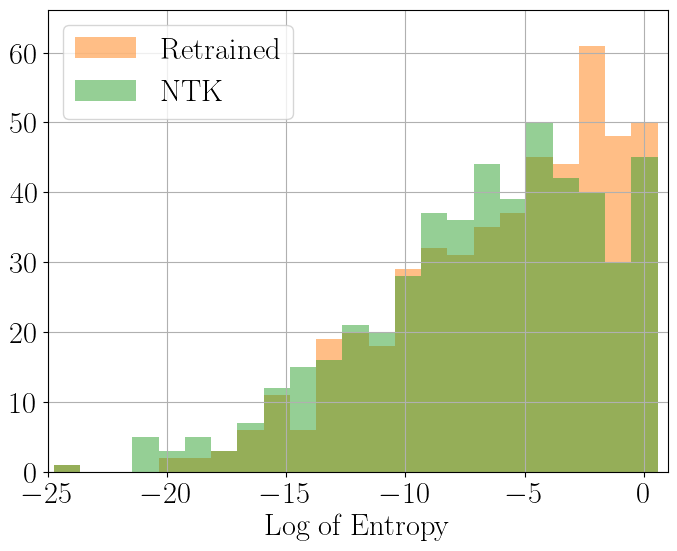}
% \caption{NTK}
\end{subfigure}
\begin{subfigure}[b]{0.24\textwidth}
\centering
\includegraphics[width=\textwidth]{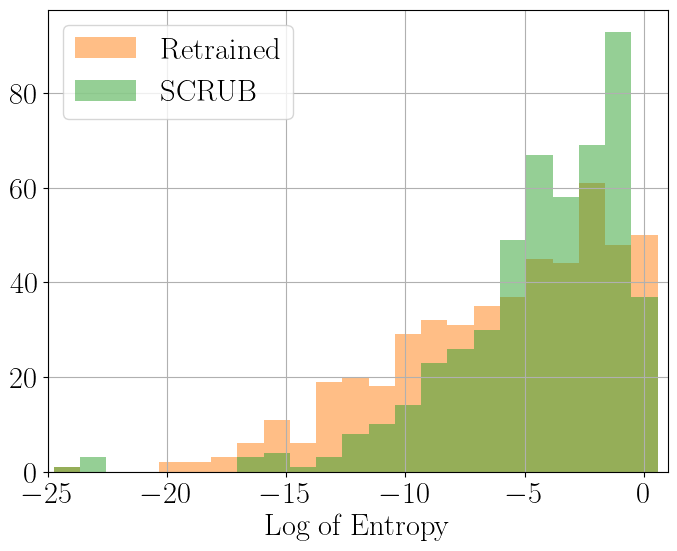}
% \caption{SCRUB}
\end{subfigure}
\begin{subfigure}[b]{0.24\textwidth}
\centering
\includegraphics[width=\textwidth]{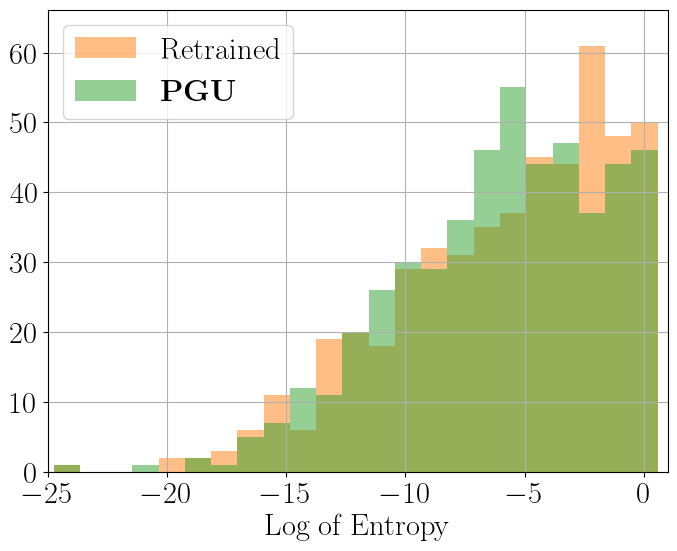}
% \caption{PGU}
\end{subfigure}
\end{subfigure}
\vspace{-1.5em}
\caption{Distribution of the entropy of model output (confidence) of forgetting dataset $\dset_f$ on original (before unlearning), retrained and unlearnt models using various methods.} 
\label{fig:allcnn_entropy}
\vspace{-1em}
\end{figure*}

\begin{figure}
% \vspace{-1.5em}
\centering
\includegraphics[width=0.3\textwidth]{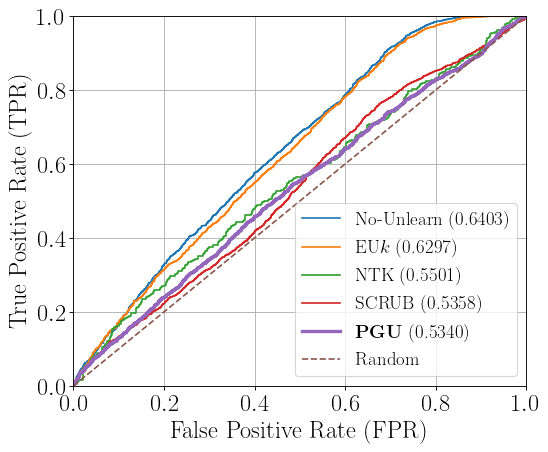}
\vspace{-1em}
\caption{MIA ROC curve for various unlearnt models to unlearn 500 samples of CIFAR10 Class 0 using AllCNN model. The number in parenthesis is AUC.} 
\label{fig:data_removal_MIA}
\vspace{-1em}
\end{figure}

%% file: tex/experiments.tex
\section{Experiments}
\subsection{Experiment setups}
\vspace{-0.5em}
\input{tex/exp_class_removal_table}
% \vspace{-0.5em}
To evaluate the effectiveness of our proposed method on forgetting samples, we conduct experiments on two different datasets: CIFAR-10 \cite{cifar10} using AllCNN \cite{allcnn}, SmallVGG (a small variation of VGG model \cite{VGG} of 3 convolutional layers and 2 linear layers\footnote{More details about this model can be found in the source code.}) and TinyImageNet \cite{tinyimagenet} using ResNet-18 \cite{resnet}. The standard CIFAR-10/TinyImageNet consists of 50K/100K for training and 10K/10K for testing. In this paper, we split the standard 10K testing set into 5K-image validation set and 5K-image testing set. We train the models using Nesterov SGD for SmallVGG, AllCNN and Adam for ResNet-18 with starting learning rate at $0.01$ and $0.001$ respectively. During training, we utilize mini-batch size of $250$ and adopt the exponential learning rate scheduler with the end learning rate set to $0.0005$ after $200$ epochs. We apply basic augmentation techniques, including random cropping and horizontal flipping. 

For the unlearning process, we update only the model weights of convolutional and linear layers, and not the weights of batch-normalization layers or biases. We adopt the exponential learning rate scheduler with the starting learning rate of $0.05$ and the ending learning rate of $0.01$ after $100$ epochs. For SmallVGG trained on CIFAR-10, we empirically choose the threshold $\gamma^l=0.95$; for AllCNN trained on CIFAR-10, we empirically choose the threshold $\gamma^l=0.9$;
and for ResNet18 trained on TinyImageNet, we empirically choose the threshold $\gamma^l=0.9$. We set $\lambda=0.2$ for all experiments.
% We also ensure that $k\ge 0.9d$, where $d$ is the input feature dimension, to avoid too small CGS.

We use the following read-out functions, which may be used to gauge how much information they were able to destroy: \textbf{(i) Error on the test set $\dset_{\text{test}}$}: ideally small, 
\textbf{(ii) Error on the subset to be forgotten $\dset_f$}: ideally the same as the error of a model trained without seeing $\dset_f$,
\textbf{(iii) Error on the retaining set $\dset_r$}: ideally no change after the unlearn process or similar to that of the retrained model,
\textbf{(iv) Membership inference attack (MIA):} To conduct a membership inference attack, we train 2 sets of models: 50 original models with full training data and 50 retrained models with only the retaining dataset. We then treat the attack as a binary classification problem on model outputs of the forget dataset, where class \textbf{1} represents the samples seen during training (i.e., from original models) and class \textbf{0} represents the samples not seen during training (i.e., from the retrained model). For this classification, we use 40 models to extract a training set, 5 models to extract a validation set, and 5 models to extract a test set. We train a XGBoost classifier and finetune to achieve the best F1 score on the validation set. We visualize the Receiver Operating Characteristic (ROC) curve and report the Area Under Curve (AUC).
% in log scale as suggested in  \cite{carlini2022membership}. 
The ROC curve shows trade-off between true-positive rate (TPR) and false-positive rate (FPR)\footnote{Please refer to Supplementary for more details of the MIA setting.}.
Furthermore, it is worth noting that our assumption about the knowledge of attacker is quite strong; specifically, we assume that the attacker knows the model architecture, training methods, the full labelled training and forgetting dataset. Employing such a strong assumption simplifies the task of quantifying the extent to which information from the forgetting set still retains within the model. This might be more challenging in a more realistic scenario due to the biases and variations caused by shallow models, shallow training-testing sets, etc.
\textbf{(v) Model confidence}: we visualize the distribution of model confidence (entropy of the output prediction) on the forget set $\dset_f$, 
\textbf{(vi) Unlearning time:} should be significantly smaller than the retraining time. 

To obtain reliable results, we conduct each experiment $5$ times and report the mean and standard deviation, except for MIA experiment which is executed only one. 
% It is important to note that applying recent studies \cite{L-CODEC,selective_forgetting,NTK_forget} on large models and datasets is not a trivial task. Simple unlearning methods, such as fine-tuning, negative gradient and random labeling, are clearly insufficient \cite{selective_forgetting}.  Furthermore, using a setting with a small model and dataset size would be not practical for our gradient-based learning method, as it it tailored to deep models and large datasets. Hence, we primarily compare our unlearned models with retrained models, which are the optimal targets for unlearning. 
We compare our method PGU with recent works: NTK\footnote{
% Similar to the setting in the original paper \cite{selective_forgetting}, we pretrain the model on CIFAR100. Additionally, 
Due to the scalability issue of the method, we can only select 1500 samples as the retaining dataset in calculation.} \cite{selective_forgetting}, Exact Unlearning-$k$ (EU-$k$)\footnote{We retrain the last 3 convolutional/linear layers. } \cite{EUk}, and SCRUB \cite{SCRUB}. 
Importantly, we note that our method only requires the forgetting dataset and is applicable even when the training data is no longer accessible.
The following experiment results demonstrate that our unlearnt method can achieve favorable results across various readout functions in comparison with other works which requires retaining dataset. Furthermore, our unlearnt models are closely matched with retrained models, which are the optimal targets for unlearning.

\input{tex/exp_data_removal}
\input{tex/exp_class_removal}
\input{tex/exp_incremental_data_removal}

\input{tex/exp_depoisoning}

%% file: tex/exp_class_removal_table.tex
% \begin{table*}[t]
% \caption{The experiment results for various readout functions for unlearning classes using ResNet-18 model trained on TinyImageNet.}
% \label{tb:class_accuracy}
% \def\arraystretch{1.1}
% % \resizebox{1.0\textwidth}{!}
% % \vspace{-1pt}
% \centering
% {
% \begin{threeparttable}
% \begin{tabular}{|l|l|c|c|c|}
% \hline
% & {Forget Set} & 5 Classes & 10 Classes & 15 Classes \\ \hline \hline
% \multirow{4}{*}{Retrained} & $\dset_{\text{retain test}}$$^\ast$ error & 53.78$\pm$0.35 & 53.58$\pm$0.45 & 53.87$\pm$0.38  \\
% & $\dset_{r}$ error  & 0.29$\pm$0.04 & 0.24$\pm$0.03 & 0.33$\pm$0.04  \\ 
% & Time (s) & 3,846$\pm$118 & 3,772$\pm$119 & 3,678$\pm$101 \\ \hline
% % & $\dset_{f}$ error & 9.10 & 9.40 & 9.00 & 9.60 & 9.60 \\ 
% % & Res Grad Norm & 0.1323 & 0.1207 & 0.1173 & 0.1273 & 0.1204 \\ \hline
% \multirow{4}{*}{\textbf{Unlearned}} & $\dset_{\text{retain test}}$$^\ast$ error & 55.14$\pm$0.55 & 55.87$\pm$0.61 & 56.53$\pm$0.60 \\
% & $\dset_{r}$ error  & 1.95$\pm$0.15 & 3.09$\pm$0.18 & 4.89$\pm$0.34  \\
% & Time (s) & 913$\pm$85 & 1,217$\pm$111 & 1,308$\pm$124 \\  \hline
% % & Res Grad Norm & 0.1446 & 0.1535 & 0.1601  & 0.18142 & 0.1796 \\ \hline
% % \multirow{2}{*}{Finetune} & $\dset_{\text{test}}$ error & x.xx & x.xx & x.xx  \\
% % & $\dset_{r}$ error & x.xx & x.xx & x.xx  \\ \hline
% % & $\dset_{f}$ error & x.xx & x.xx & x.xx & x.xx & x.xx \\ 
% \end{tabular}
% \begin{tablenotes}
% \footnotesize
% \item $^\ast$ $\dset_\text{test}$ after excluding samples of forgetting classes.
% \end{tablenotes}
% \end{threeparttable}
% }
% \end{table*}

\begin{table*}[t]
\caption{The experiment results for various readout functions for unlearning classes using ResNet-18 model trained on TinyImageNet.}

\label{tb:class_accuracy}
\def\arraystretch{1.1}
% \resizebox{1.0\textwidth}{!}
\vspace{-0.75em}
\centering
{
\begin{threeparttable}
\begin{tabular}{|l|l|c|c|c|c|c|c|}
\hline
 & Method & Original & Retrained & EU$k$ & SCRUB & \textbf{PGU} \\ \hline \hline
\multirow{3}{*}{5 Classes} & $\dset_{\text{retain test}}$$^\ast$ error & 52.78$\pm$0.36 & 52.58$\pm$0.17 & 53.15$\pm$0.28 & 51.80$\pm$0.37 & 53.03$\pm$0.43 \\
& $\dset_{r}$ error & 0.56$\pm$0.04 & 0.56$\pm$0.05 & 1.84$\pm$0.08 & 0.20$\pm$0.00 & 0.74$\pm$0.06 \\ 
& Time (s) & -- & 3,846$\pm$118 & 2318$\pm$97 & 1693$\pm$65 & 765$\pm$45 \\ \hline

% \multirow{3}{*}{{10 Classes}} & $\dset_{\text{retain test}}$$^\ast$ error & 53.58$\pm$0.45 & 53.37$\pm$0.32 & 53.08$\pm$0.32 & 53.75$\pm$0.61\\
% & $\dset_{r}$ error & 0.24$\pm$0.03 & 0.28$\pm$0.03 & 0.15$\pm$0.01 & 1.19$\pm$0.18  \\
% & Time (s) & 3,772$\pm$119 & 2224$\pm$103 & 1605$\pm$59 & 873$\pm$71 \\  \hline
%
% \multirow{3}{*}{{15 Classes}} & $\dset_{\text{retain test}}$$^\ast$ error & 53.87$\pm$0.38 & x.xx$\pm$x.xx & x.xx$\pm$x.xx & 56.53$\pm$0.60 \\
% & $\dset_{r}$ error  & 0.33$\pm$0.04 & x.xx$\pm$x.xx & x.xx$\pm$x.xx & 4.89$\pm$0.34  \\
% & Time (s) & 3,678$\pm$101 & x.xx$\pm$x.xx & x.xx$\pm$x.xx & 1,308$\pm$124 \\  \hline
\end{tabular}
\begin{tablenotes}
\footnotesize
\item $^\ast$ $\dset_\text{test}$ after excluding samples of forgetting classes.
\end{tablenotes}
\end{threeparttable}
}
\vspace{-1em}
\end{table*}

\begin{figure*}[t]
\centering
\begin{subfigure}[b]{\textwidth}
\begin{subfigure}[b]{0.24\textwidth}
\centering
\includegraphics[width=\textwidth]{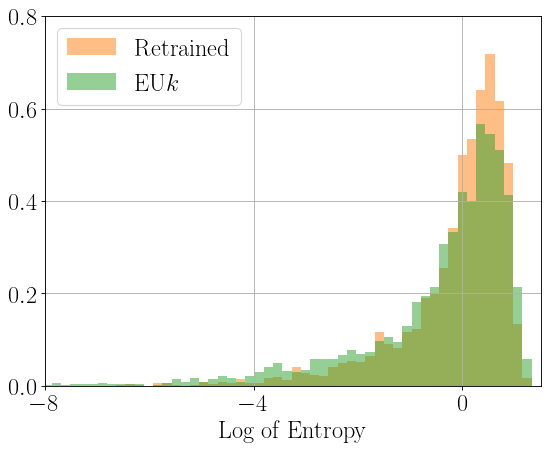}
\caption{}
\end{subfigure}
\begin{subfigure}[b]{0.24\textwidth}
\centering
\includegraphics[width=\textwidth]{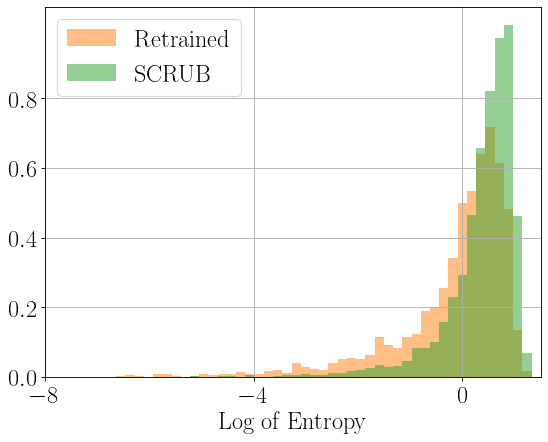}
\caption{}
\end{subfigure}
\begin{subfigure}[b]{0.24\textwidth}
\centering
\includegraphics[width=\textwidth]{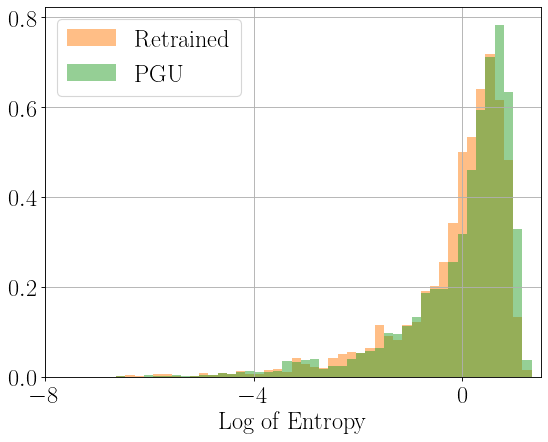}
\caption{}
\end{subfigure}
\begin{subfigure}[b]{0.24\textwidth}
\centering
\includegraphics[width=\textwidth]{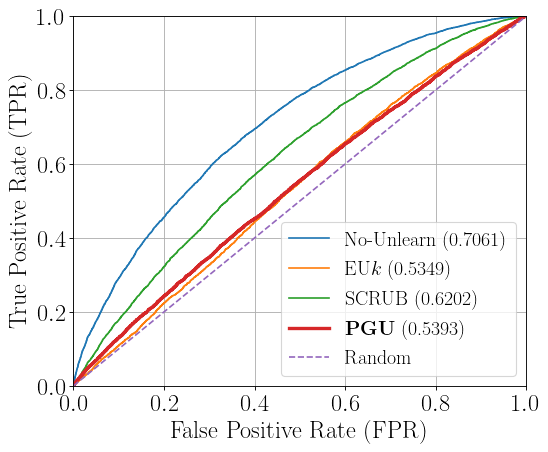}
\caption{}
\end{subfigure}
\end{subfigure}
\vspace{-1.5em}
\caption{Distribution of the entropy of model output (confidence) of forgetting dataset $\dset_f$ on retrained and unlearnt models using various methods (Fig a, b, and c) and MIA ROC curve (Fig d) when forgetting 5 classes of TinyImageNet using ResNet-18.} 
\label{fig:resnet_entropy}
\vspace{-1em}
\end{figure*}

%% file: tex/exp_data_removal.tex
\subsection{Data removal}
First, we conduct experiment to selective remove some training data. Specifically, we remove $500$ samples of the first class (Class 0) of CIFAR10 dataset using AllCNN.

The experimental results, as presented in Table \ref{tb:compare}, show the favorable performance of our proposed method in comparison to alternative approaches. Remarkably, our unlearnt model closely matches with the retrained model even without using the retaining dataset. Specifically, our method exhibits the smallest increase in error on $\dset_{\text{test}}$. Notably, we found that EU$k$ method \cite{EUk} performs poorly on many readout functions, i.e., $\dset_{f}$ error is very small and similar to $\dset_{r}$ error. This observation suggests that even though the last few layers are retrained from scratch, information from the forgetting dataset can still be leaked from shallower layers. We observe a larger increases in $\dset_{r}$ $\dset_{\text{test}}$ errors for NTK methods; this is potentially because we can only sample 1500 samples (out of 49,500 samples) as the retaining dataset due to its scalability limitation. 

Considering model confidence, as depicted in Figure \ref{fig:allcnn_entropy}, it is evident that our unlearnt method produces a $\dset_{f}$ entropy distribution which aligns with that of the retrained model the best.
NTK, despite of its scalability limitation, can produce a $\dset_{f}$ entropy distribution which also closely aligns with that of retrained model.
Furthermore, the $\dset_{f}$ entropy distribution of EU$k$ is closer the the $\dset_{f}$ entropy distribution of original model than that of retrain model. This outcome corroborates the findings from Table \ref{tb:compare}. 

Figure \ref{fig:data_removal_MIA} demonstrates the effectiveness of our approach in erasing information related to forgotten datasets from the model. Specifically, there is a substantial decrease in the success rate of MIA attackers, with the Area Under Curve (AUC) dropping from $0.6403$ to $0.5340$ (slightly higher than random chances).
In term of AUC for MIA ROC curve, our method achieves comparable results to SCRUB \cite{SCRUB}, showcasing its merits as it does not require the retaining dataset. Finally, 
% our method is faster than the retrained approach and other methods. Our method is faster than other approaches since we only requires the forgetting dataset, which is usually small, while other methods need to fine-tunning on the retaining dataset. We note that the training time of our method does include the SVD computation time. 
our approach outpaces both the retrained method and alternative techniques in terms of speed. Its efficiency comes from the fact that it solely relies on the forgetting dataset, usually at a small size; whereas alternative methods need to fine-tuning on the large retaining dataset. It is worth mentioning that the training time of our approach includes the SVD computation time, specifically SVD computation consumes less than 6 seconds in this experiment.

%% file: tex/exp_class_removal.tex
\subsection{Class removals}
% \vspace{-0.5em}

% \begin{figure}[t]
% \centering
% \begin{subfigure}[b]{0.235\textwidth}
% \includegraphics[width=\textwidth]{figs/resnet18_tinyimagenet/MIA_5classes1.png}
% \caption{5 Classes}
% \end{subfigure}
% \begin{subfigure}[b]{0.235\textwidth}
% \includegraphics[width=\textwidth]{figs/resnet18_tinyimagenet/MIA_10classes.png}
% \caption{10 Classes}
% % \caption{}
% % \label{fig:class_removal_forget_dist}
% \end{subfigure}
% \caption{MIA ROC curve of forgetting 5 and 10 classes using ResNet-18 trained on TinyImageNet.
% } 
% \label{fig:class_removal}
% \vspace{-1em}
% \end{figure}

% \begin{figure}
% % \vspace{-1.5em}
% \centering
% \includegraphics[width=0.4\textwidth]{figs/resnet18_tinyimagenet/MIA_5classes2.png}
% \vspace{-1em}
% \caption{MIA ROC curve of forgetting 5 classes using ResNet-18 trained on TinyImageNet.} 
% \label{fig:class_removal_cm}
% \vspace{-1.25em}
% \end{figure}

\begin{figure}
% \vspace{-1.5em}
\centering
\includegraphics[width=0.35\textwidth]{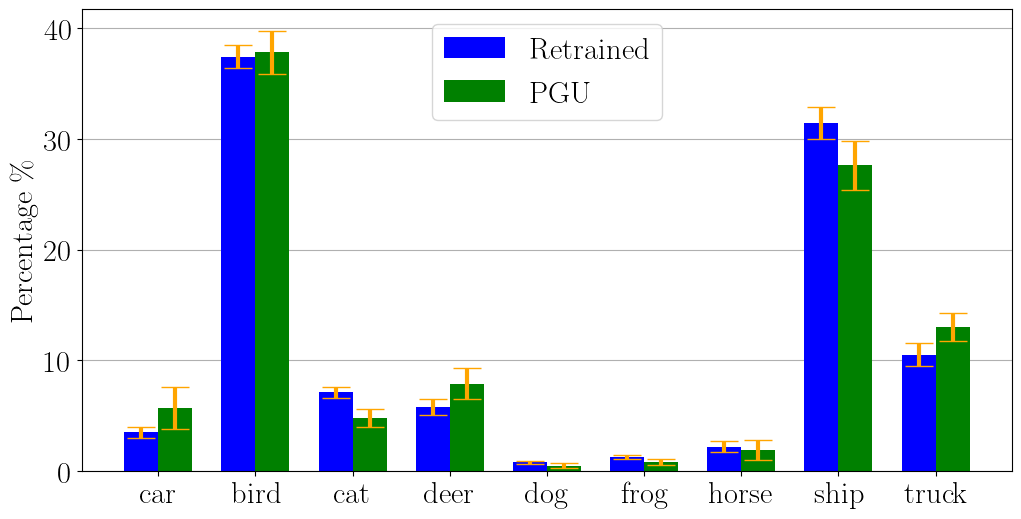}
\vspace{-1em}
\caption{How retrained and our unlearnt models classify forgetting samples (airplane).} 
\label{fig:class_removal_cm}
\vspace{-1.2em}
\end{figure}

\input{tex/exp_incremental_data_removal_table}

The above setting shows applications where samples are randomly removed. 
Another appealing application of unlearning involves completely removing samples from specific classes. To test the effectiveness of our proposed method in this application, we conducted experiments using ResNet-18 on TinyImageNet dataset and SmallVGG on CIFAR-10 dataset. Without losing generality, we select first 5 classes of TinyImageNet and the first class of CIFAR-10 as the forgetting set.  Noting that for this class removal task, we remove the rows corresponding to forgetting classes in the last linear layer of unlearnt models before evaluating.

In Table \ref{tb:class_accuracy}, we present the outcomes of our experiments with different readout functions. We observe that classification errors of the our unlearnt model on retaining training and retaining testing set (composed of the training and testing sets excluding the samples from the forgetting class) are more similar to those of the retrained models as compared to EU$k$, SCRUB. The $\dset_{f}$ entropy distribution of our unlearnt model also aligns with that of the retrained model better than EU$k$, SCRUB (Figure \ref{fig:resnet_entropy}a, b, c).
In term of MIA when forgetting 5 classes, our method can achieve better AUC than that of SCRUB while only being slight lower than that of EU$k$ (Figure \ref{fig:resnet_entropy}d). Finally, our method is faster than other methods and significantly faster than the retrained approach. In summary, our comprehensive analysis across various metrics underscores the advantages offered by our proposed method.

We further analyse the impact of unlearning a class on retaining classes using SmallVGG on CIFAR-10 and unlearning the first class (i.e., airplane).
Figure \ref{fig:class_removal_cm} depicts how (percentage \% of samples) the unlearnt and retrained models will classify the forgotten samples in CIFAR-10 dataset. The figure shows that the unlearnt and retrained models exhibit similar behaviors when presented airplane samples. Specifically, we can see that both models will more likely to classify an airplane sample as bird or ship, potentially due to the similarity in the background of those classes. Conversely, both models are also less likely to classify an airplane as a dog, a frog, or a horse.

%% file: tex/exp_incremental_data_removal_table.tex
\begin{table*}[t]
\caption{The experiment results for various readout functions when conducting incremental unlearning on SmallVGG trained on CIFAR10.}
\label{tb:accuracy}
\def\arraystretch{1.1}
% \resizebox{1.0\textwidth}{!}
\vspace{-0.5em}
\centering
{
\begin{threeparttable}
\begin{tabular}{|l|l|c|c|c|c|}
\hline
& {Num. Forget} & 0 - 1K & 1K - 2K & 2K - 3K & 3K - 4K \\ \hline \hline
\multirow{4}{*}{Retrained} & $\dset_{\text{test}}$ error (\%)  & 9.62$\pm$0.52 & 9.55$\pm$0.44 & 9.38$\pm$0.32 & 9.84$\pm$0.43 \\
& $\dset_{r}$ error (\%)   & 0.00$\pm$0.00 & 0.00$\pm$0.00 & 0.00$\pm$0.00 & 0.00$\pm$0.00 \\ 
& $\dset_{f}$ error (\%)  & 9.61$\pm$0.53 & 9.51$\pm$0.43 & 9.59$\pm$0.30 & 9.69$\pm$0.26 \\
% & $\dset_{f}$ error (\%)  & 9.10 & 9.40 & 9.00 & 9.60  \\ % \hline
& Time (s) & 1030$\pm$19 & 1020$\pm$20 & 1009$\pm$21 & 995$\pm$22 \\ \hline
\multirow{4}{*}{\textbf{PGU}} & $\dset_{\text{test}}$ error (\%)  & 9.84$\pm$1.08 & 9.86$\pm$0.99 & 9.78$\pm$0.95 & 9.80$\pm$1.11 \\
& $\dset_{r}$ error (\%)   & 0.00$\pm$0.00 & 0.00$\pm$0.00 & 0.01$\pm$0.00 & 0.01$\pm$0.00 \\ 
& $\dset_{f}$ error (\%)  & 10.02$\pm$0.97 & 9.93$\pm$1.10 & 10.02$\pm$1.24 & 11.30$\pm$1.37 \\
& Time (s) & 156$\pm$21 & 161$\pm$19 & 178$\pm$29  & 172$\pm$24 \\ \hline
\end{tabular}
% \begin{tablenotes}
% \footnotesize
% \item $^\dag$ We note that, for a retrained model, 
% \end{tablenotes}
\end{threeparttable}
}
\vspace{-0.5em}
\end{table*}

\begin{figure*}[t]
\centering
\begin{subfigure}[b]{\textwidth}
\begin{subfigure}[b]{0.24\textwidth}
\centering
\includegraphics[width=\textwidth]{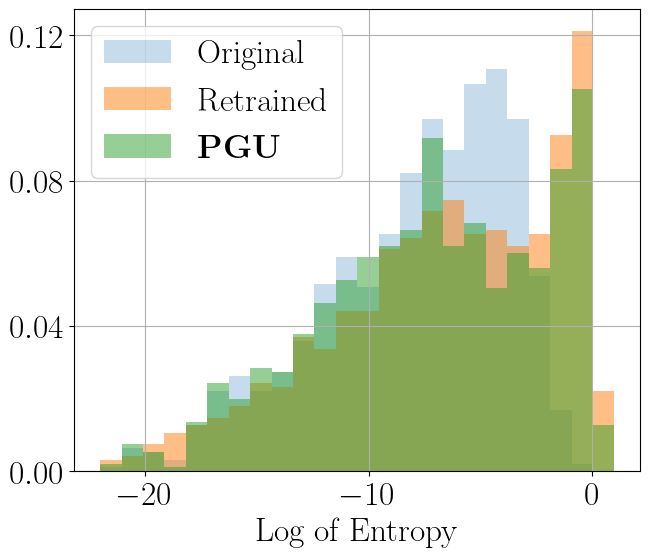}
\caption{0 - 1K}
\end{subfigure}
\begin{subfigure}[b]{0.24\textwidth}
\centering
\includegraphics[width=\textwidth]{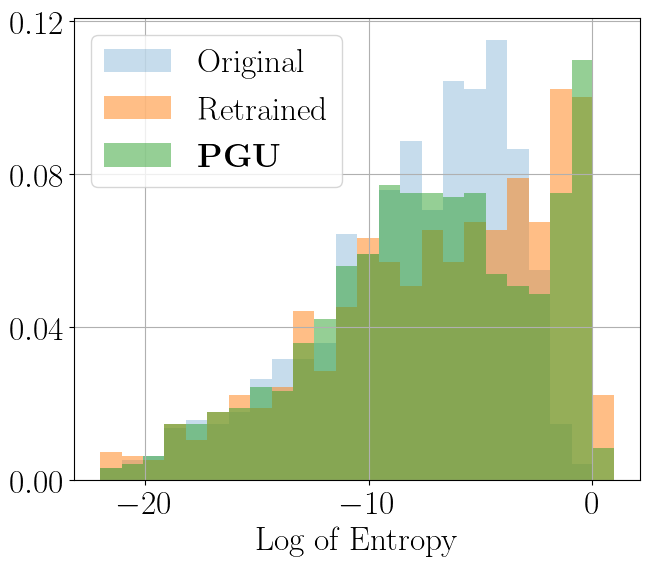}
\caption{1K - 2K}
\end{subfigure}
\begin{subfigure}[b]{0.24\textwidth}
\centering
\includegraphics[width=\textwidth]{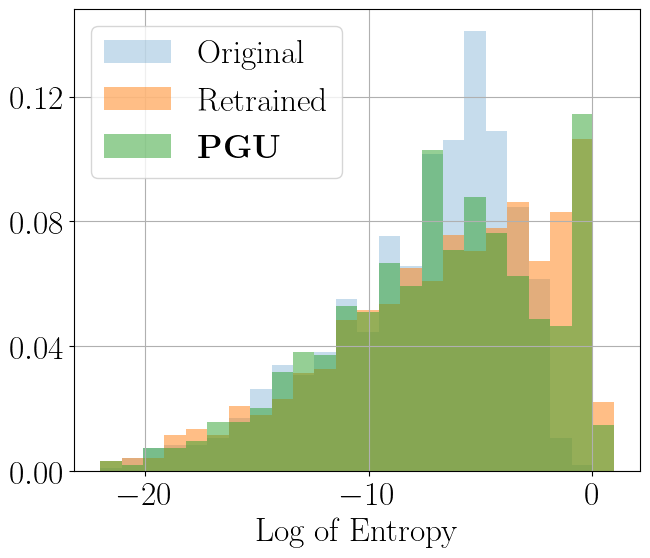}
\caption{2K - 3K}
\end{subfigure}
\begin{subfigure}[b]{0.24\textwidth}
\centering
\includegraphics[width=\textwidth]{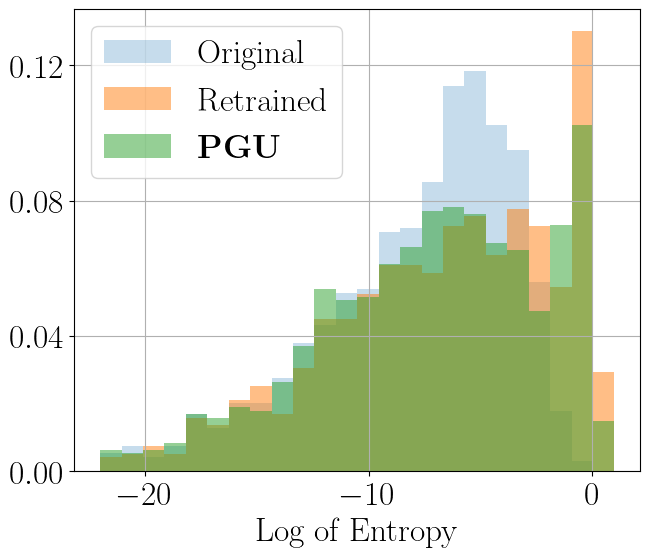}
\caption{3K - 4K}
\end{subfigure}
\end{subfigure}
\vspace{-1.5em}
\caption{Distribution of the entropy of model output (confidence) of forgetting dataset $\dset_f$ on original (before unlearning), retrained and our unlearnt models.} 
\label{fig:entropy}
\vspace{-1.2em}
\end{figure*}

\begin{figure}
% \vspace{-1.5em}
\centering
\includegraphics[width=0.28\textwidth]{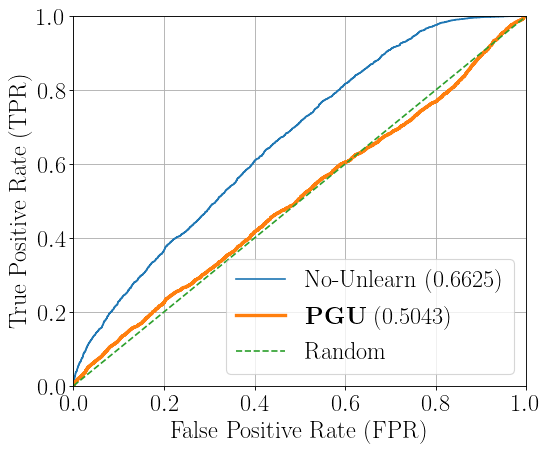}
\vspace{-1em}
\caption{MIA ROC curve after incremental unlearning 4K samples (1K step size) of CIFAR10 on SmallVGG model.} 
\label{fig:MIA_ROC_smallvgg}
\vspace{-0.5em}
\end{figure}

%% file: tex/exp_incremental_data_removal.tex
\input{tex/exp_depoisoning_table}

\subsection{{Incremental data removals}}
We performed sequential unlearning experiments where we unlearnt a subset of 1,000 training samples at a time. In the first run, we unlearnt a set of 1,000 samples, and the resulting unlearnt model was used to subsequently unlearn another subset of 1,000 training samples. The subset of 1,000 forget samples in each run are formed by randomly selecting 100 samples for each class in the remaining training samples. 

Table \ref{tb:accuracy} demonstrates that the unlearnt model achieves very similar test errors to the retrained models, while only minor increases in error rate for retaining set $\dset_r$ and forgetting set $\dset_f$ are observed.
In terms of MIA success rate, Figure \ref{fig:MIA_ROC_smallvgg} shows that our method results in the ROC curve that closely aligns with the random line. This suggests that attackers are unable to determine whether a forgotten sample was employed in training process any more effectively than random guessing.

Additionally, Figure \ref{fig:entropy} shows the distributions of log-entropy of the unlearnt and retrained models on the forgetting dataset $\dset_f$ before and after unlearning. We can observe that after unlearning the log-entropy distributions of the unlearnt and retrained models are closely matched to each other. Figure \ref{fig:entropy} and Table \ref{tb:accuracy} indicate that our method can support incremental unlearning without significantly affecting performance of test set $\dset_{\text{test}}$. Finally, our method is significantly faster than the retrained approach.

%% file: tex/exp_depoisoning_table.tex
\begin{table}[t]
\caption{Test accuracy for different poisoning budgets for SmallVGG trained CIFAR10. The SmallVGG model achieves 90.16\% in test accuracy with \textit{full-clean} training data. We present the accuracy of the model trained with poisoned dataset (\textbf{\textit{Poisoned}}), the model trained with clean data only (exclude the poisoned samples) (\textbf{\textit{Clean}}), and the {poisoned} model after depoisoning  using PGU and \cite{unlearn_features_labels}.}
\label{tb:depoisoning}
\def\arraystretch{1.1}
% \resizebox{1.0\textwidth}{!}
\vspace{-0.25em}
\centering
{
\begin{tabular}{|l|c|c|c|c|c|}
\hline
Num. Poison & 5K & 10K & 15K & 20K & 25K \\ \hline \hline
Poisoned    & 86.07 & 79.32 & 69.69 & 56.78 & 45.36 \\
Clean  & 89.81 & 88.89 & 88.69 & 87.68 & 86.86 \\ \hline
\textbf{PGU}  & \textbf{87.95} & \textbf{86.87} & \textbf{85.46} & \textbf{85.13} & \textbf{83.12} \\ \hline
1st Order \cite{unlearn_features_labels}  & 86.56 & 83.77 & 79.68 & 74.11 & 67.98 \\ 
2nd Order \cite{unlearn_features_labels}  & 87.43 & 84.13 & 79.80 & 74.28 & 68.28 \\ \hline
\end{tabular}
}
\vspace{-0.75em}
\end{table}

%% file: tex/exp_depoisoning.tex
\subsection{Depoisoning}
In this section, we focus on the poisoning attack setting on machine learning. Specifically, an adversary aims at misleading a model by flipping labels of the training data. The label flips significantly degrade the performance of the learning model. Hence, we use the unlearning method to remove the harmful effect of poisoned samples and correct the model to achieve the performance as if the model was trained without the poisoned samples.  

% For this depoisoning application, since we not only want the model to unlearn the features (could be noise) that is correlated with the "poisoned" labels, but we also expect the model can re-classify those poisoned samples correctly. Hence, we instead use $\lambda < 0$ to minimize the entropy of the model outputs (i.e., we want the model gives a high confident score on another class which is different from the poisoned labels). 
% For this de-poisoning application, we not only want the model to unlearn the features (which could be noise) that are correlated with the "poisoned" labels, but we also expect the model to correctly re-classify those poisoned samples. Therefore, we use $\lambda < 0$ to minimize the entropy of the model outputs. In other words, we want the model to give a high confidence score to a different class that is different from the poisoned labels.

For this experiment, we use the SmallVGG model on CIFAR10 dataset. 
For fair comparison to \cite{unlearn_features_labels}, we follow their experimental setting. Specifically, we change the labels of certain pairs of classes by flipping a portion of them to their counterpart. For example, a sample with the label "cat" could be changed to "dog" and vice versa. This type of attack can result in similar performance degradation as other label-flip attacks \cite{unlearn_features_labels}. We evaluate our unlearning method on this setting with different poisoning budgets.

% \begin{figure}[t]
% \centering

% \begin{subfigure}[b]{0.49\textwidth}
% \centering
% \includegraphics[width=0.6\textwidth]{figs/depoisoning_1000.png}
% \caption{10,000 poisoned samples}
% \end{subfigure}
% \begin{subfigure}[b]{0.49\textwidth}
% \centering

% \includegraphics[width=0.6\textwidth]{figs/depoisoning_2000.png}
% \caption{20,000 poisoned samples}
% \end{subfigure}
% \caption{Accuracy of the depoisoned model on various datasets during the training process. The \textit{"Clean"} dataset includes only clean training samples, while the \textit{"Poison"} dataset contains only poisoned training samples. The \textit{"Poison w. Clean label"} dataset is composed of poisoned training samples evaluated with clean labels.} 
% \label{fig:depoisoning}
% \end{figure}

The experimental results are presented in Table \ref{tb:depoisoning}. These results demonstrate that our unlearning method can effectively mitigate the detrimental effects of poisoned samples on the model, resulting in significant performance recovery across different levels of poisoning (including cases where up to 50\% of the samples are poisoned). Additionally, our method also significantly surpasses the method in \cite{unlearn_features_labels} in term of performance recovery, especially when a large number of poisoned samples are presented.
Importantly, our approach does not require access to the clean labels of the poisoned samples.
% Figure \ref{fig:depoisoning} illustrates that the test accuracy can be quickly restored within a few epochs of training. However, we note that training for more epochs may degrade the model's performance, as discussed in Section \ref{sec:incremental_unlearning}.

%% file: tex/conclusion.tex
\section{Conclusion}
\vspace{-0.5em}
We introduce a new machine unlearning method to remove the effect of a specific subset of training data on the trained model. It has important applications in ensuring the “right to be forgotten” in the context of user privacy, erasing a subset of malicious or adversarial data from the model. The proposed method shows promising empirical performance for different model architectures and datasets across various readout functions. We also show the ability to approximately unlearn for large models very efficiently without any additional limitations beyond those encountered during training. Additionally, as not requiring the retaining dataset, our proposed method can be effectively employed when the training data is no longer accessible, a scenario where the majority of existing approaches are not applicable. For future work, we plan to extend this approach to other learning applications beyond classification.
% such as recommendation system, segmentation, NLP; and also investigating the machine unlearning problem for unsupervised/self-supervised and generative scenarios. 
Additionally, we aim to test the effectiveness of our method against advanced Membership Inference Attack and model inversion techniques, as these are active research areas.
% we plan to explore the proposed method to other learning application beyond classification,
% Furthermore, we remark that Membership Inference Attack (MIA) and model inversion are active research fields, we look forward to testing our method against more advanced MIA and model inversion methods to confirm the effectiveness of our proposed method.

%% file: tex/ablation_studies.tex
\subsection{Ablation studies}

\subsubsection{Effect of $\gamma^l$}

To assess the impact of $\gamma^l$ on the efficacy of our unlearning approach, we conducted experiments involving the removal of 1,000 samples from the CIFAR-10 dataset (comprising 100 samples per class) using the SmallVGG model. In Table \ref{tb:by_var}, we present the classification error rates for the forgotten dataset $\dset_{f}$ and the retained dataset $\dset_{r}$ as we manipulate the value of $\gamma^l$.

Our empirical findings indicate that our method remains robust over a wide range of $\gamma^l$ values.
With a large $\gamma^l$ (approximately 1, i.e., 0.99 or 0.995), the updating space, i.e., the Residual Gradient Space (RGS), is exceedingly small, hindering the unlearning process. Consequently, the model struggles to eliminate the information associated with the forgotten dataset (lower classification error on forgetting dataset).
Conversely, when $\gamma^l$ is too small ($< 0.8$), the updating space becomes expansive and contains substantial information about the retainning dataset, this results in a drop in performance on both the retained dataset and the test dataset undesirably. Furthermore, we note that as the updating space is larger, the unlearning process is easier to be over-fit, which could result in the Streisand Effect.
An appropriately $\gamma^l$ facilitates the unlearning process while preserving the information of the retained dataset intact.

\begin{table}[t]
\caption{Classification error of the forgetting $\dset_{f}$ and retaining $\dset_r$ dataset when forgetting 1000 samples of CIFAR-10 dataset using SmallVGG model at different values of $\gamma^l$.}
\label{tb:by_var}
\def\arraystretch{1.1}
\centering
\begin{tabular}{|c|l|c|c|c|}
\hline 
 % & \multicolumn{2}{c|}{AUC} \\ \hline  
&  & \multicolumn{3}{c|}{Classification Error} \\ \hline  
&  & $\dset_{f}$ & $\dset_r$ & $\dset_{\text{test}}$ \\ \hline \hline 
\multicolumn{2}{|c|}{Retrain} & 9.61 & 0.00 & 9.62 \\ \hline
\multirow{8}{*}{$\gamma^l$} & 0.995 & 1.10 & 0.00 & 9.90 \\
& 0.99 & 4.28 & 0.00 & 9.88 \\
& 0.98 & 10.04 & 0.00 & 9.86 \\ 
& 0.96 & 10.06 & 0.00 & 9.84 \\ 
& 0.90 & 10.06 & 0.00 & 9.84 \\ 
& 0.85 & 10.04 & 0.00 & 9.88 \\
& 0.80 & 13.80 & 1.74 & 13.78 \\
& 0.75 & 19.30 & 8.15 & 19.14 \\\hline 
\end{tabular}
\end{table}
\vspace{-1em}

\subsubsection{Layers}
In this section, we conduct experiments to assess the efficacy of our proposed method when only unlearning specific layers.
In a manner akin to CF$k$ \cite{EUk}, we freeze the first $k$ layers (the shallower layers), while we exclusively update the deeper layers. 
More specifically, we conduct two set of experiments: (i) SmallVGG model on CIFAR10 dataset to forget 100 samples per class; and (ii) ResNet-18 model on TinyImageNet dataset to forget the first 5 classes.
Experimental results in Table \ref{tb:by_layer} and Table \ref{tb:by_layer_resnet18} reveal that it is possible eliminate information of forgetting dataset retaining in the model outputs by unlearning only a subset of layers instead of the whole models. 
Yet, the number of layers which needs to update is depended on the architecture and dataset. We will further investigate this problem in future works.
% However, unlearning only the last linear layer is not enough to discard information of the forgetting data. 
In comparison to EU$k$, simply retraining the last few layers  does not effectively remove information of the forgetting dataset on the data removal task. While on the class removal task, retraining the last few layers can be a good approach.

\begin{table}[t]
\caption{AUC of the MIA ROC with different numbers of unlearned layers of SmallVGG model for GPU and EU$k$.}
\label{tb:by_layer}
\def\arraystretch{1.1}
\centering
\begin{tabular}{|l|c|c|}
\hline 
 % & \multicolumn{2}{c|}{AUC} \\ \hline  
Updating top-$k$ & PGU & EU$k$ \\ \hline \hline 
0 layer (No-Unlearn) & \multicolumn{2}{c|}{0.6590} \\ \hline  
1 layer & 0.6554 & 0.6587 \\
2 layers & 0.5086 & 0.5587 \\
3 layers & 0.5087 & 0.5379 \\ 
4 layers & 0.5067 & 0.5168 \\ 
5 layers (full model) & 0.5070 & 0.5040 \\ \hline 
\end{tabular}
\end{table}
\vspace{-1em}

\begin{table}[t]
\caption{AUC of the MIA ROC with different numbers of unlearned layer blocks of ResNet-18 model for GPU and EU$k$. For simplicity, we consider the first convolutional layer and the last linear layer as a block.}
\label{tb:by_layer_resnet18}
\def\arraystretch{1.1}
\centering
\begin{tabular}{|l|c|c|}
\hline 
 % & \multicolumn{2}{c|}{AUC} \\ \hline  
Updating top-$k$ & PGU & EU$k$ \\ \hline \hline 
0 block (No-Unlearn) & \multicolumn{2}{c|}{0.7053} \\ \hline  
1 block & 0.7035 & 0.5350 \\
2 blocks & 0.6063 & 0.5349 \\
3 blocks & 0.5409 & 0.5243 \\ 
4 blocks & 0.5395 & 0.5198 \\ 
5 blocks & 0.5393 & 0.5045 \\
6 blocks (full model) & 0.5392 & 0.5045 \\ \hline 
\end{tabular}
\vspace{-1em}
\end{table}

\subsubsection{Accuracy losses vs. AUC of MIA ROC}
\begin{figure}[t]
% \vspace{-1.5em}
\centering
\includegraphics[width=0.35\textwidth]{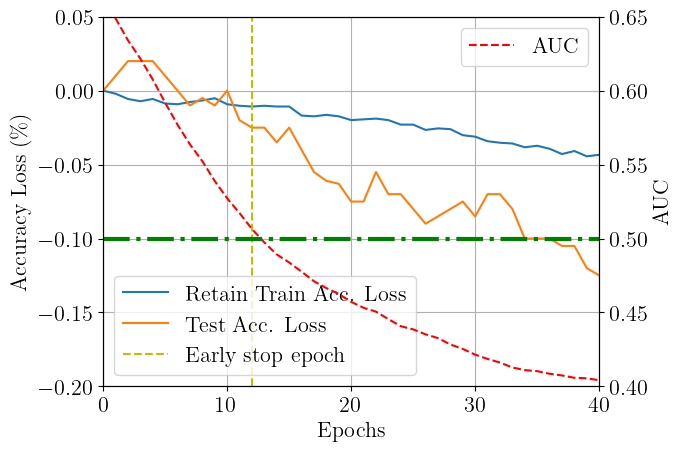}
\vspace{-1em}
\caption{Accuracy loses (compared to the original model) of retaining training and testing dataset (left y-axis) and AUC of MIA ROC curves (right y-axis) by epoch when unlearning 100 smaples per class CIFAR-10 using SmallVGG.} 
\label{fig:class_removal_acc_vs_MIA_smallvgg}
\vspace{-0.75em}
\end{figure}

\begin{figure}[t]
% \vspace{-1.5em}
\centering
\includegraphics[width=0.35\textwidth]{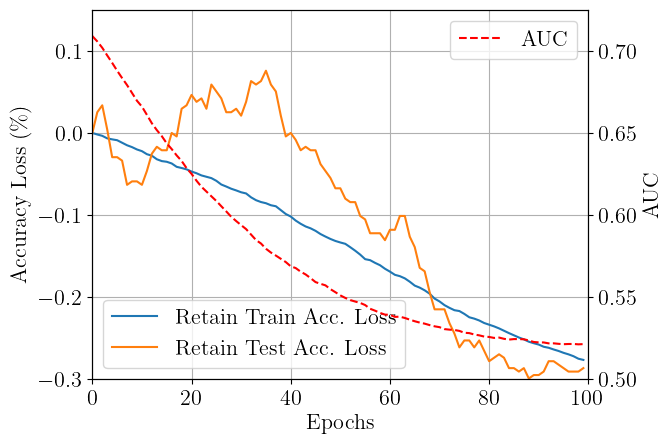}
\vspace{-1em}
\caption{Accuracy loses (compared to the original model) of retaining training and testing dataset (left y-axis) and AUC of MIA ROC curves (right y-axis) by epoch when unlearning 5 classes of TinyImageNet using ResNet-18.} 
\label{fig:class_removal_acc_vs_MIA}
\vspace{-0.75em}
\end{figure}

\begin{figure*}[t]
\centering
\begin{subfigure}[b]{\textwidth}
\begin{subfigure}[b]{0.32\textwidth}
\centering
\includegraphics[width=\textwidth]{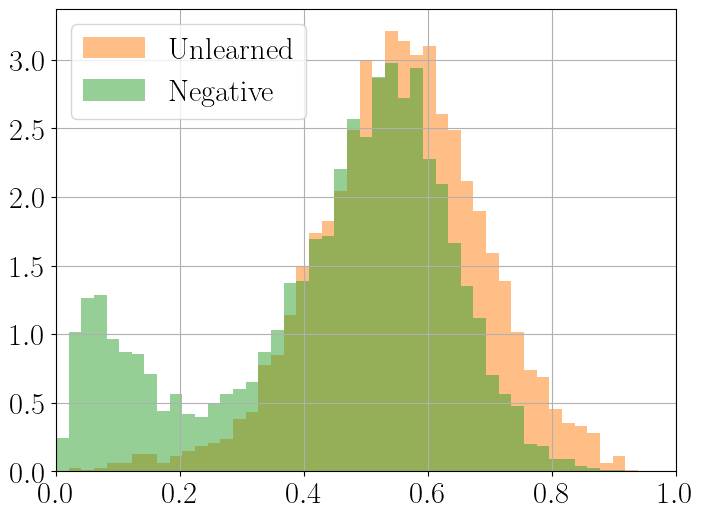}
\caption{Epoch 0 (Original)}
\end{subfigure}
~~
\begin{subfigure}[b]{0.32\textwidth}
\centering
\includegraphics[width=\textwidth]{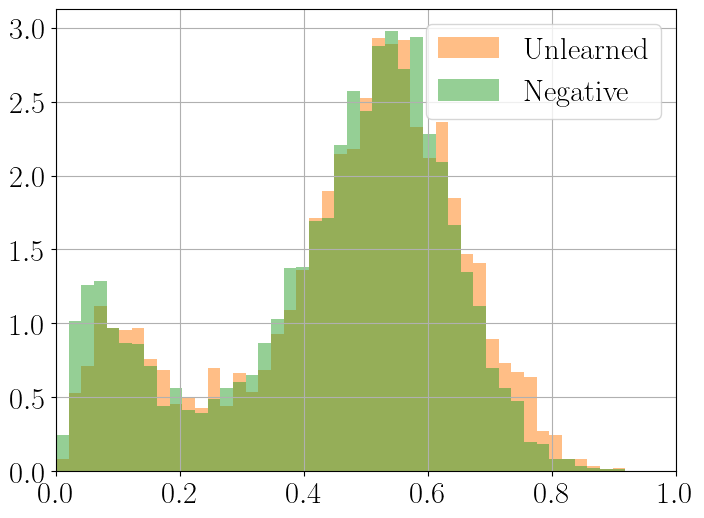}
\caption{Epoch 12 (Early stop point)}
\end{subfigure}
~~
\begin{subfigure}[b]{0.32\textwidth}
\centering
\includegraphics[width=\textwidth]{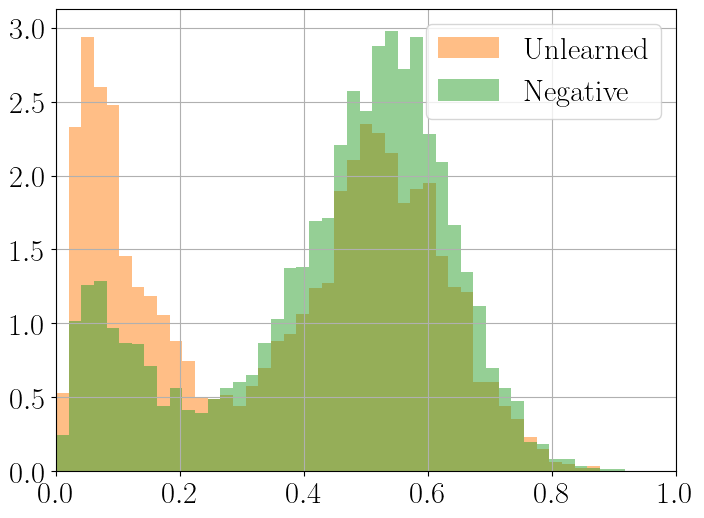}
\caption{Epoch 40}
\end{subfigure}
\end{subfigure}
\vspace{-1.5em}
\caption{Distributions of the MI attacker confidence scores of unlearned testing set (comprising unlearned and negative samples) during unlearning 100 samples per class of CIFAR-10 using SmallVGG model at different time points: epoch 0 (original), epoch 12 (early stop point) and epoch 40.} 
\label{fig:entropy_by_epoch}
\vspace{-1em}
\end{figure*}

We also analyse the trade-off between accuracy losses in retaining training/testing dataset and the AUC of MIA ROC curve. 
For the data removal task, from Figure \ref{fig:class_removal_acc_vs_MIA_smallvgg} and Figure \ref{fig:entropy_by_epoch}, we can observe that training for more epochs could result in over-fitting for unlearning, which could result in the Streisand Effect. 
More specifically, Figure \ref{fig:entropy_by_epoch}c shows that when unlearned model is over-fitted, a large number of unlearned samples have very small MIA confidence scores. This could be exploited by attackers to determine membership.
Hence, it is important to early stop to avoid over-fitting.
For the class-removal task, since we remove the rows corresponding to forgetting classes in the last linear layer of unlearned models, we do not observe the Streisand Effect as in data-removal task. The AUC of ROC curve does not go lower than random level 0.5. 
% Additionally, the distribution of MIA confidence scores at the end of unlearning process also shows that unlearned samples does not h
Therefore, we can achieve lower AUC of MIA ROC curve at the cost of more losses in the accuracy of the retaining training and testing dataset (and training time). However, it's noteworthy that the AUC remains relatively stable after a certain number of training epochs, e.g., the 70th epoch in Figure \ref{fig:class_removal_acc_vs_MIA}, indicating that prolonging training beyond this point does not yield discernible benefits. 
% The training time of our proposed method increases almost linearly with regard to the number of epochs.
In both experiments, we see that the performance drops of retaining dataset and testing dataset are quite small.

\begin{figure*}[t]
\centering
\begin{subfigure}[b]{\textwidth}
\begin{subfigure}[b]{0.32\textwidth}
\centering
\includegraphics[width=\textwidth]{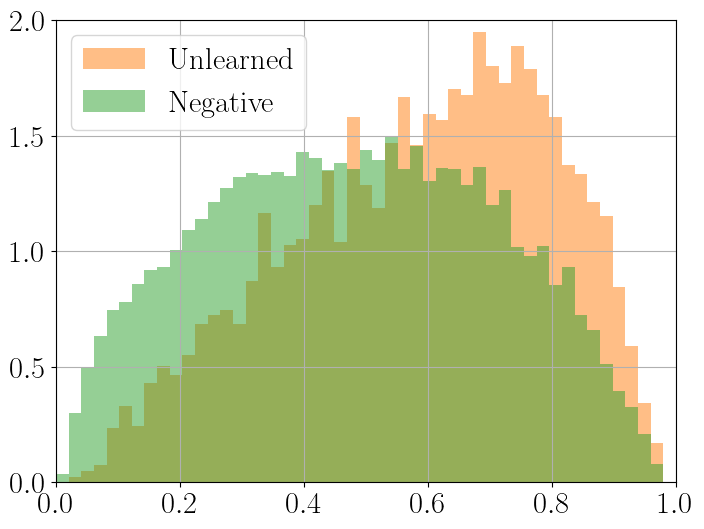}
\caption{Epoch 0 (Original)}
\end{subfigure}
~~
\begin{subfigure}[b]{0.32\textwidth}
\centering
\includegraphics[width=\textwidth]{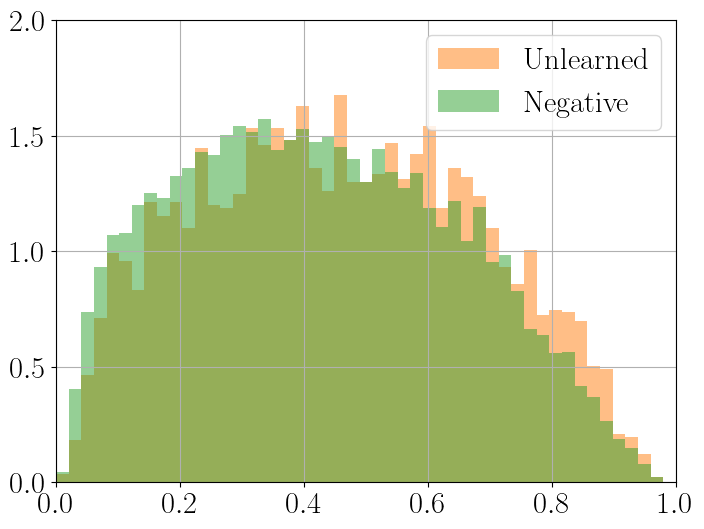}
\caption{Epoch 70}
\end{subfigure}
~~
\begin{subfigure}[b]{0.32\textwidth}
\centering
\includegraphics[width=\textwidth]{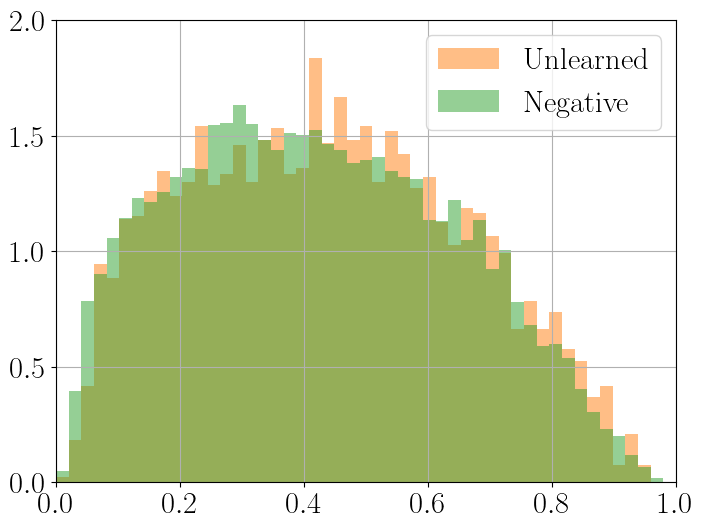}
\caption{Epoch 99}
\end{subfigure}
\end{subfigure}
\vspace{-1.5em}
\caption{Distributions of the MI attacker confidence scores of unlearned testing set (comprising unlearned and negative samples) during unlearning the first 5-classes of TinyImangeNet dataset using ResNet-18 model at different time points: epoch 0 (original), epoch 70 and epoch 99.} 
\label{fig:entropy_by_epoch_resnet}
\vspace{-1em}
\end{figure*}

\subsubsection{Incremental learning limitation}
\begin{figure}[t]
% \vspace{-1.5em}
\centering
\includegraphics[width=0.35\textwidth]{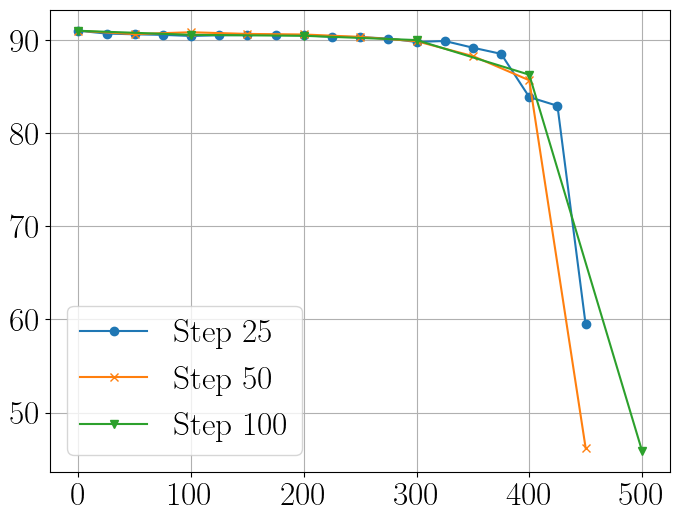}
\vspace{-1em}
\caption{Testing set accuracy of incremental unlearning with step sizes of 25, 50, or 100 samples per class of CIFAR-10 using AllCNN.} 
\label{fig:allcnn_incremental}
\vspace{-0.75em}
\end{figure}

% In this sub-section, we will conduct stress tests to analyse unlearning method will fail (i.e., large accuracy drops on test set) after how many unlearning steps. We conduct experiments using AllCNN and SmallVGG using CIFAR-10 dataset with the data removal task. We incrementally unlearn 25, 50, or 100 samples per classes using AllCNN model and incrementally unlearn 100 samples per class using SmallVGG model. 
In this section, we conduct stress tests to analyze the point at which our unlearning method fails, characterized by significant accuracy drops on the test set. We conduct experiments using AllCNN and SmallVGG on CIFAR-10 dataset with the data removal task. We incrementally unlearn 25, 50, or 100 samples per class with the AllCNN model, and 100 samples per class with the SmallVGG model.
Figure \ref{fig:allcnn_incremental} shows that regardless of step sizes, AllCNN model starts to fail after unlearning 400 samples per class (8\% of training size); while SmallVGG model starts to fail at 1100 samples per class (22\% of training size). The SmallVGG model fails at much larger number of forgetting samples than AllCNN does potentially because SmallVGG model has more parameters (5.7M) than AllCNN dose (1.6M). This allows SmallVGG to be easily updated to remove  information of forgetting dataset. Nevertheless, in both models, we believe that the forgetting dataset sizes are large enough for the life cycle of a model and it is reasonable to retrain the model. 
Finally, we note that we can easily recover the testing accuracy drops by fine-tuning the unlearned model on retaining dataset; however, in this paper we mainly focus on the setting that the training dataset may be no longer accessible.

\begin{figure}[t]
% \vspace{-1.5em}
\centering
\includegraphics[width=0.35\textwidth]{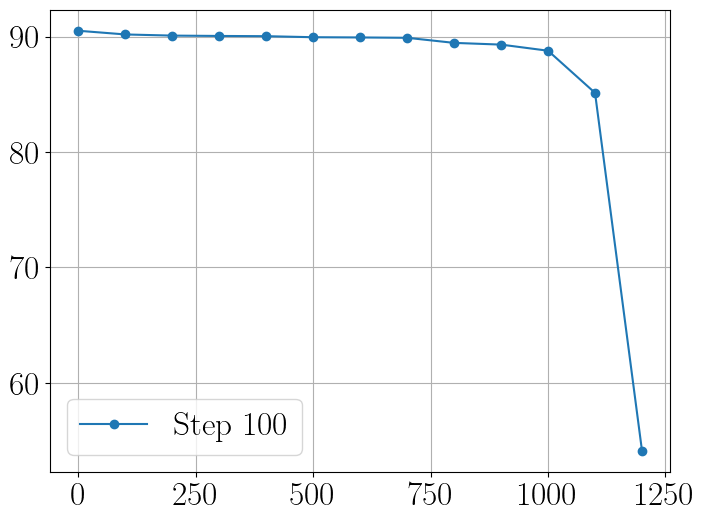}
\vspace{-1em}
\caption{Testing set accuracy of incremental unlearn with step size of 100 samples per class of CIFAR-10 using SmallVGG.} 
\label{fig:smallvgg_incremental}
% \vspace{-0.75em}
\end{figure}

\subsection{Depoisoning}
We conduct further experiments on the depoisoning setting using ResNet-18 model on TinyImageNet dataset. The experimental results in Table \ref{tb:depoisoning_1} again confirm effectiveness of our method on mitigating the detrimental effects of poisoned samples on a larger model and dataset. Our method can even achieve better test performance than the model that is trained on the clean dataset only when a large number of training dataset is poisoned. 

\begin{table}[t]
\caption{Test accuracy for different poisoning budgets for ResNet-18 trained TinyImageNet. The ResNet-18 model achieves 48.18\% in test accuracy with \textit{full-clean} training data. We present the accuracy of the model trained with poisoned dataset (\textbf{\textit{Poisoned}}), the model trained with clean data only (exclude the poisoned samples) (\textbf{\textit{Clean}}), and the {poisoned} model after depoisoning using PGU.
% and \cite{unlearn_features_labels}.
}
\label{tb:depoisoning_1}
\def\arraystretch{1.1}
% \resizebox{1.0\textwidth}{!}
\vspace{-0.25em}
\centering
{
\begin{tabular}{|l|c|c|c|c|c|}
\hline
Num. Poison & 10K & 20K & 30K & 40K & 50K \\ \hline \hline
Poisoned & 44.06 & 39.76 & 34.42 & 28.70 & 22.40 \\
Clean    & 46.40 & 45.30 & 43.90 & 42.34 & 39.48 \\ \hline
\textbf{PGU}  & \textbf{45.38} & \textbf{44.34} & \textbf{43.60} & \textbf{42.50} & \textbf{41.14} \\ \hline
% 1st Order \cite{unlearn_features_labels}  & 86.56 & 83.77 & 79.68 & 74.11 & 67.98 \\ 
% 2nd Order \cite{unlearn_features_labels}  & 87.43 & 84.13 & 79.80 & 74.28 & 68.28 \\ \hline
\end{tabular}
}
\vspace{-0.5em}
\end{table}

%% file: PaperForReview.bbl
\begin{thebibliography}{10}\itemsep=-1pt

\bibitem{selective_forgetting}
{Aditya Golatkar and Alessandro Achille and Stefano Soatto}.
\newblock {Eternal Sunshine of the Spotless Net: Selective Forgetting in Deep
  Networks}.
\newblock In {\em CVPR}, pages 9301--9309, 2020.

\bibitem{andrew2021differentially}
Galen Andrew, Om Thakkar, Hugh~Brendan McMahan, and Swaroop Ramaswamy.
\newblock Differentially private learning with adaptive clipping.
\newblock In A. Beygelzimer, Y. Dauphin, P. Liang, and J.~Wortman Vaughan,
  editors, {\em Advances in Neural Information Processing Systems}, 2021.

\bibitem{ml_linear_filtration}
Thomas Baumhauer, Pascal Sch\"{o}ttle, and Matthias Zeppelzauer.
\newblock Machine unlearning: Linear filtration for logit-based classifiers.
\newblock 111(9):3203–3226, 2022.

\bibitem{SISA}
Lucas Bourtoule, Varun Chandrasekaran, Christopher~A. Choquette-Choo, Hengrui
  Jia, Adelin Travers, Baiwu Zhang, David Lie, and Nicolas Papernot.
\newblock Machine unlearning.
\newblock In {\em 2021 IEEE Symposium on Security and Privacy (SP)}, pages
  141--159, 2021.

\bibitem{7163042}
Yinzhi Cao and Junfeng Yang.
\newblock Towards making systems forget with machine unlearning.
\newblock In {\em 2015 IEEE Symposium on Security and Privacy}, pages 463--480,
  2015.

\bibitem{carlini2022membership}
Nicholas Carlini, Steve Chien, Milad Nasr, Shuang Song, Andreas Terzis, and
  Florian Tramer.
\newblock Membership inference attacks from first principles.
\newblock In {\em 2022 IEEE Symposium on Security and Privacy (SP)}, pages
  1897--1914. IEEE, 2022.

\bibitem{236216}
Nicholas Carlini, Chang Liu, {\'U}lfar Erlingsson, Jernej Kos, and Dawn Song.
\newblock The secret sharer: Evaluating and testing unintended memorization in
  neural networks.
\newblock In {\em 28th USENIX Security Symposium (USENIX Security 19)}, pages
  267--284, Aug. 2019.

\bibitem{choquette2021label}
Christopher~A Choquette-Choo, Florian Tramer, Nicholas Carlini, and Nicolas
  Papernot.
\newblock Label-only membership inference attacks.
\newblock In {\em International conference on machine learning}, pages
  1964--1974, 2021.

\bibitem{private_data_analysis}
Cynthia Dwork, Frank McSherry, Kobbi Nissim, and Adam Smith.
\newblock Calibrating noise to sensitivity in private data analysis.
\newblock In Shai Halevi and Tal Rabin, editors, {\em Theory of Cryptography},
  pages 265--284, 2006.

\bibitem{Algorithmic_DP}
Cynthia Dwork and Aaron Roth.
\newblock The algorithmic foundations of differential privacy.
\newblock {\em Found. Trends Theor. Comput. Sci.}, 9(3–4):211–407, aug
  2014.

\bibitem{10.1145/2810103.2813677}
Matt Fredrikson, Somesh Jha, and Thomas Ristenpart.
\newblock Model inversion attacks that exploit confidence information and basic
  countermeasures.
\newblock In {\em Proceedings of the 22nd ACM SIGSAC Conference on Computer and
  Communications Security}, CCS '15, page 1322–1333, 2015.

\bibitem{fu2022knowledge}
Shaopeng Fu, Fengxiang He, and Dacheng Tao.
\newblock Knowledge removal in sampling-based bayesian inference.
\newblock In {\em International Conference on Learning Representations}, 2022.

\bibitem{EUk}
Shashwat Goel, Ameya Prabhu, and Ponnurangam Kumaraguru.
\newblock Evaluating inexact unlearning requires revisiting forgetting, 01
  2022.

\bibitem{9577384}
A. Golatkar, A. Achille, A. Ravichandran, M. Polito, and S. Soatto.
\newblock Mixed-privacy forgetting in deep networks.
\newblock In {\em 2021 IEEE/CVF Conference on Computer Vision and Pattern
  Recognition (CVPR)}, pages 792--801, jun 2021.

\bibitem{NTK_forget}
Aditya Golatkar, Alessandro Achille, and Stefano Soatto.
\newblock {Forgetting Outside the Box: Scrubbing Deep Networks of Information
  Accessible from Input-Output Observations}.
\newblock In Andrea Vedaldi, Horst Bischof, Thomas Brox, and Jan-Michael Frahm,
  editors, {\em Computer Vision -- ECCV 2020}, pages 383--398, 2020.

\bibitem{Golatkar_2022_CVPR}
Aditya Golatkar, Alessandro Achille, Yu-Xiang Wang, Aaron Roth, Michael Kearns,
  and Stefano Soatto.
\newblock Mixed differential privacy in computer vision.
\newblock In {\em Proceedings of the IEEE/CVF Conference on Computer Vision and
  Pattern Recognition (CVPR)}, pages 8376--8386, June 2022.

\bibitem{gopi2021numerical}
Sivakanth Gopi, Yin~Tat Lee, and Lukas Wutschitz.
\newblock Numerical composition of differential privacy.
\newblock In {\em NeurIPS 2021}, June 2021.

\bibitem{certified_linear_removal}
Chuan Guo, Tom Goldstein, Awni Hannun, and Laurens Van Der~Maaten.
\newblock Certified data removal from machine learning models.
\newblock In {\em Proceedings of the 37th International Conference on Machine
  Learning}, ICML'20, 2020.

\bibitem{resnet}
Kaiming He, Xiangyu Zhang, Shaoqing Ren, and Jian Sun.
\newblock {Deep Residual Learning for Image Recognition}.
\newblock In {\em CVPR}, 2016.

\bibitem{pmlr-v130-izzo21a}
Zachary Izzo, Mary Anne~Smart, Kamalika Chaudhuri, and James Zou.
\newblock Approximate data deletion from machine learning models.
\newblock In {\em Proceedings of The 24th International Conference on
  Artificial Intelligence and Statistics}, volume 130 of {\em Proceedings of
  Machine Learning Research}, pages 2008--2016, 13--15 Apr 2021.

\bibitem{5484614}
Masayuki Karasuyama and Ichiro Takeuchi.
\newblock Multiple incremental decremental learning of support vector machines.
\newblock {\em IEEE Transactions on Neural Networks}, 21(7):1048--1059, 2010.

\bibitem{9857498}
Junyaup Kim and Simon~S. Woo.
\newblock Efficient two-stage model retraining for machine unlearning.
\newblock In {\em 2022 IEEE/CVF Conference on Computer Vision and Pattern
  Recognition Workshops (CVPRW)}, pages 4360--4368, 2022.

\bibitem{influence_function}
Pang~Wei Koh and Percy Liang.
\newblock Understanding black-box predictions via influence functions.
\newblock In {\em Proceedings of the 34th International Conference on Machine
  Learning - Volume 70}, ICML'17, page 1885–1894, 2017.

\bibitem{cifar10}
Alex Krizhevsky.
\newblock Learning multiple layers of features from tiny images.
\newblock Technical report, University of Toronto, 2009.

\bibitem{SCRUB}
Meghdad Kurmanji, Peter Triantafillou, and Eleni Triantafillou.
\newblock Towards unbounded machine unlearning, 02 2023.

\bibitem{NTK}
Jaehoon Lee, Lechao Xiao, Samuel~S. Schoenholz, Yasaman Bahri, Roman Novak,
  Jascha Sohl-Dickstein, and Jeffrey Pennington.
\newblock Wide neural networks of any depth evolve as linear models under
  gradient descent.
\newblock In {\em Proceedings of the 33rd International Conference on Neural
  Information Processing Systems}, 2019.

\bibitem{trgp}
Sen Lin, Li Yang, Deliang Fan, and Junshan Zhang.
\newblock {TRGP: Trust Region Gradient Projection for Continual Learning}.
\newblock 2022.

\bibitem{GDPR}
A. Mantelero.
\newblock {The EU proposal for a general data protection regulation and the
  roots of the ‘right to be forgotten’}.
\newblock In {\em Computer Law and Security Review}, page 29(3):229–235,
  2013.

\bibitem{10.1145/3457607}
Ninareh Mehrabi, Fred Morstatter, Nripsuta Saxena, Kristina Lerman, and Aram
  Galstyan.
\newblock A survey on bias and fairness in machine learning.
\newblock {\em ACM Comput. Surv.}, 54(6), jul 2021.

\bibitem{L-CODEC}
Ronak Mehta, Sourav Pal, Vikas Singh, and Sathya~N. Ravi.
\newblock Deep unlearning via randomized conditionally independent hessians.
\newblock In {\em Proceedings of the IEEE/CVF Conference on Computer Vision and
  Pattern Recognition (CVPR)}, pages 10422--10431, June 2022.

\bibitem{melis2019exploiting}
Luca Melis, Congzheng Song, Emiliano De~Cristofaro, and Vitaly Shmatikov.
\newblock Exploiting unintended feature leakage in collaborative learning.
\newblock In {\em 2019 IEEE symposium on security and privacy (SP)}, pages
  691--706. IEEE, 2019.

\bibitem{phong_2}
Quoc~Phong Nguyen, Bryan Kian, Hsiang Low, and Patrick Jaillet.
\newblock Variational bayesian unlearning.
\newblock In {\em Proceedings of the 34th International Conference on Neural
  Information Processing Systems}, NIPS'20, 2020.

\bibitem{phong_1}
Quoc~Phong Nguyen, Ryutaro Oikawa, Dinil~Mon Divakaran, Mun~Choon Chan, and
  Bryan Kian~Hsiang Low.
\newblock Markov chain monte carlo-based machine unlearning: Unlearning what
  needs to be forgotten.
\newblock In {\em Proceedings of the 2022 ACM on Asia Conference on Computer
  and Communications Security}, ASIA CCS '22, page 351–363, 2022.

\bibitem{CCPA}
Stuart~L. Pardau.
\newblock {THE CALIFORNIA CONSUMER PRIVACY ACT: TOWARDS A EUROPEAN-STYLE
  PRIVACY REGIME IN THE UNITED STATES?}
\newblock In {\em Journal of Technology Law and Policy}, volume~23, 2018.

\bibitem{10.1007/978-3-540-74690-4_22}
Enrique Romero, Ignacio Barrio, and Llu{\'i}s Belanche.
\newblock Incremental and decremental learning for linear support vector
  machines.
\newblock In Joaquim~Marques de S{\'a}, Lu{\'i}s~A. Alexandre, W{\l}odzis{\l}aw
  Duch, and Danilo Mandic, editors, {\em Artificial Neural Networks -- ICANN
  2007}, pages 209--218, 2007.

\bibitem{white_black_MI}
Alexandre Sablayrolles, Matthijs Douze, Cordelia Schmid, Yann Ollivier, and
  Herv{\'e} J{\'e}gou.
\newblock White-box vs black-box: Bayes optimal strategies for membership
  inference.
\newblock In {\em International Conference on Machine Learning}, 2019.

\bibitem{9605653}
Gobinda Saha, Isha Garg, Aayush Ankit, and Kaushik Roy.
\newblock Space: Structured compression and sharing of representational space
  for continual learning.
\newblock {\em IEEE Access}, 9:150480--150494, 2021.

\bibitem{GPM}
Gobinda Saha, Isha Garg, and Kaushik Roy.
\newblock Gradient projection memory for continual learning.
\newblock In {\em International Conference on Learning Representations}, 2021.

\bibitem{remember_forget}
Ayush Sekhari, Jayadev Acharya, Gautam Kamath, and Ananda~Theertha Suresh.
\newblock Remember what you want to forget: Algorithms for machine unlearning.
\newblock In A. Beygelzimer, Y. Dauphin, P. Liang, and J.~Wortman Vaughan,
  editors, {\em Advances in Neural Information Processing Systems}, 2021.

\bibitem{7958568}
R. Shokri, M. Stronati, C. Song, and V. Shmatikov.
\newblock Membership inference attacks against machine learning models.
\newblock In {\em 2017 IEEE Symposium on Security and Privacy (SP)}, pages
  3--18, Los Alamitos, CA, USA, may 2017. IEEE Computer Society.

\bibitem{VGG}
Karen Simonyan and Andrew Zisserman.
\newblock Very deep convolutional networks for large-scale image recognition.
\newblock {\em arXiv preprint arXiv:1409.1556}, 2014.

\bibitem{allcnn}
Jost~Tobias Springenberg, Alexey Dosovitskiy, Thomas Brox, and Martin
  Riedmiller.
\newblock {Striving for Simplicity: The All Convolutional Net}.
\newblock In {\em ICLR Workshop}, 2015.

\bibitem{unrollingSGD}
A. Thudi, G. Deza, V. Chandrasekaran, and N. Papernot.
\newblock Unrolling sgd: Understanding factors influencing machine unlearning.
\newblock In {\em 2022 IEEE 7th European Symposium on Security and Privacy
  (EuroSP)}, pages 303--319, jun 2022.

\bibitem{unlearn_features_labels}
Alexander Warnecke, Lukas Pirch, Christian Wressnegger, and Konrad Rieck.
\newblock {Machine Unlearning of Features and Labels}.
\newblock {\em Network and Distributed System Security (NDSS)}, 2023.

\bibitem{7536387}
Xi Wu, Matthew Fredrikson, Somesh Jha, and Jeffrey~F. Naughton.
\newblock A methodology for formalizing model-inversion attacks.
\newblock In {\em 2016 IEEE 29th Computer Security Foundations Symposium
  (CSF)}, pages 355--370, 2016.

\bibitem{tinyimagenet}
Ya~Le; Xuan~S. Yang.
\newblock Tiny imagenet visual recognition challenge.
\newblock Technical report, 2015.

\bibitem{zhang2017understanding}
Chiyuan Zhang, Samy Bengio, Moritz Hardt, Benjamin Recht, and Oriol Vinyals.
\newblock {Understanding deep learning requires rethinking generalization}.
\newblock In {\em International Conference on Learning Representations}, 2017.

\bibitem{PCMU}
Zijie Zhang, Yang Zhou, Xin Zhao, Tianshi Che, and Lingjuan Lyu.
\newblock Prompt certified machine unlearning with randomized gradient
  smoothing and quantization.
\newblock In {\em Advances in Neural Information Processing Systems}, pages
  13433--13455, 2022.

\end{thebibliography}
